\documentclass[journal]{IEEEtran}

\ifCLASSOPTIONcompsoc
\usepackage[caption=false, font=normalsize, labelfont=sf, textfont=sf]{subfig}
\else
\usepackage[caption=false, font=footnotesize]{subfig}
\fi
\usepackage{amsfonts}
\usepackage{latexsym}
\usepackage{booktabs}
\usepackage{amsmath}
\ifCLASSINFOpdf
\usepackage[pdftex]{graphicx}
 \graphicspath{{./}}
\DeclareGraphicsExtensions{.pdf,.jpeg,.png}
\else
 \usepackage[dvips]{graphicx}
 \graphicspath{{../eps/}}
\DeclareGraphicsExtensions{.eps}
\fi

\usepackage{array}
\newcolumntype{I}{!{\vrule width 3pt}}
\newlength\savedwidth

\newlength\savewidth
\newcommand\shline{\noalign{\global\savewidth\arrayrulewidth
		\global\arrayrulewidth 1.5pt}%
	\hline
	\noalign{\global\arrayrulewidth\savewidth}}

\begin{document}
\title{R2RNet: Low-light Image Enhancement via Real-low to Real-normal Network}

\author{Jiang Hai,
        Zhu Xuan,
        Songchen Han,
        Ren Yang, Yutong Hao, Fengzhu Zou,
        and~Fang Lin
\thanks{This work is partially supported by Key R\&D project of Sichuan Province, China (No.22ZDYF3720). \emph{(Corresponding author: Songchen Han.)}}
\thanks{The authors are with the School of Aeronautics and Astronautics, Sichuan University, Chengdu 610065, China (e-mail: j1269998232@163.com, j\_zhxxx@163.com, renyang20212021@163.com, 15522808379@163.com, Y15286544560@163.com, photoatoms@163.com, hansongchen@scu.edu.cn).}
\thanks{}}

\markboth{Journal of \LaTeX\ Class Files,~Vol.~14, No.~8, August~2015}%
{Shell \MakeLowercase{\textit{et al.}}: Bare Demo of IEEEtran.cls for IEEE Journals}

\maketitle

\begin{abstract}
Images captured in weak illumination conditions could seriously degrade the image quality. Solving a series of degradation of low-light images can effectively improve the visual quality of images and the performance of high-level visual tasks. In this study, a novel Retinex-based Real-low to Real-normal Network (R2RNet) is proposed for low-light image enhancement, which includes three subnets: a Decom-Net, a Denoise-Net, and a Relight-Net. These three subnets are used for decomposing, denoising, contrast enhancement and detail preservation, respectively.  Our R2RNet not only uses the spatial information of the image to improve the contrast but also uses the frequency information to preserve the details. Therefore, our model acheived more robust results for all degraded images. Unlike most previous methods that were trained on synthetic images, we collected the first Large-Scale Real-World paired low/normal-light images dataset (LSRW dataset) to satisfy the training requirements and make our model have better generalization performance in real-world scenes. Extensive experiments on publicly available datasets demonstrated that our method outperforms the existing state-of-the-art methods both quantitatively and visually. In addition, our results showed that the performance of the high-level visual task (\emph{i.e.} face detection) can be effectively improved by using the enhanced results obtained by our method in low-light conditions. Our codes and the LSRW dataset are available at: https://github.com/abcdef2000/R2RNet. 
\end{abstract}


\IEEEpeerreviewmaketitle

\section{Introduction}
\IEEEPARstart{I}{NSUFFICIENT} illumination in the image capturing seriously affects the image quality from many aspects, such as low contrast and low visibility. Removing these degradations and transforming a low-light image into a high-quality sharp image is helpful to improve the performance of high-level visual tasks, such as image recognition \cite{1}, object detection \cite{2}, semantic segmentation \cite{3}, \emph{etc}, and can also improve the performance of intelligent systems in some practical applications, such as autonomous driving, visual navigation \cite{4}, \emph{etc}. Low-light image enhancement, therefore, is highly desired.

Over the past few decades, there have been a large number of methods employed to enhance degraded images captured under insufficient illumination conditions. These methods have made great progress in improving image contrast and can obtain enhanced images with better visual quality. In addition to contrast, another special degradation of low-light images is noise. Many methods utilized additional denoising methods as pre-processing or post-processing. However, using denoising methods as pre-processing will cause blurring, while applying denoising as post-processing will result in noise amplification \cite{5}. Recently, some methods \cite{6} have designed effective models to perform denoising and contrast enhancement simultaneously and obtain satisfactory results. 

\begin{figure}[!t]
	\centering
	\subfloat[Input]{
		\begin{minipage}[a]{0.115\textwidth}
			\centering
				\includegraphics[width=0.82in]{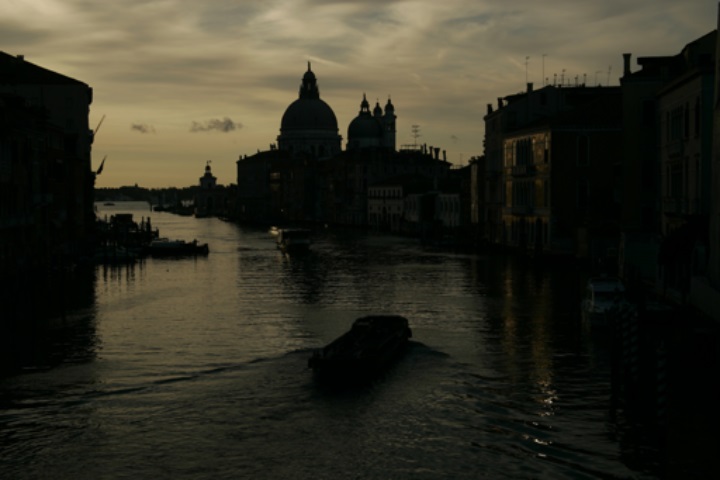}\vspace{2pt}
				\includegraphics[width=0.82in]{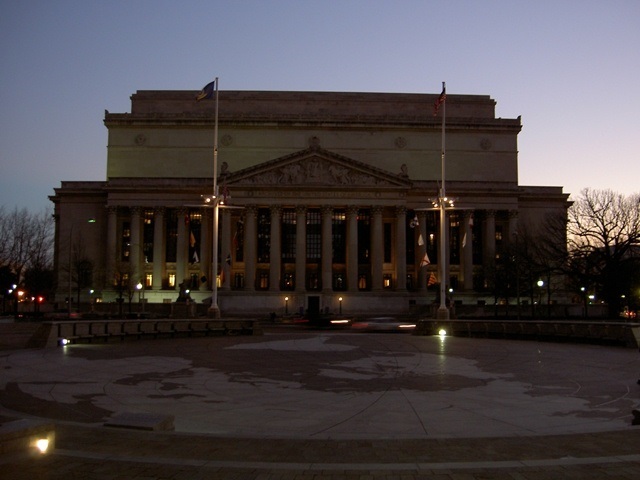}\vspace{2pt}
		\end{minipage}}
	\subfloat[SRIE]{
			\begin{minipage}[a]{0.115\textwidth}
				\centering
					\includegraphics[width=0.82in]{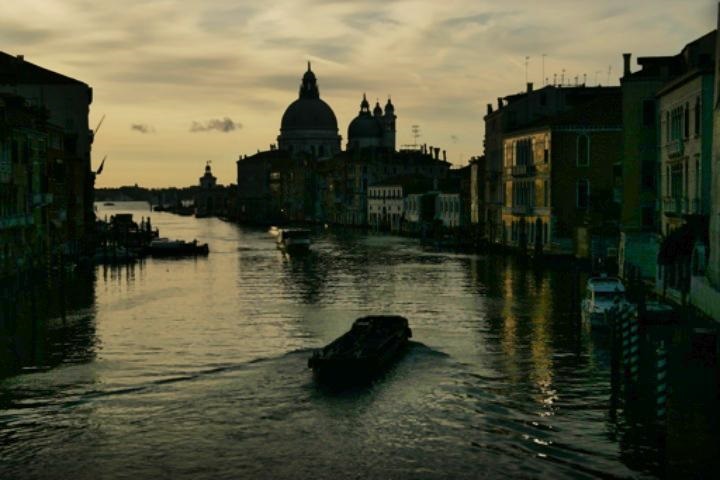}\vspace{2pt}
					\includegraphics[width=0.82in]{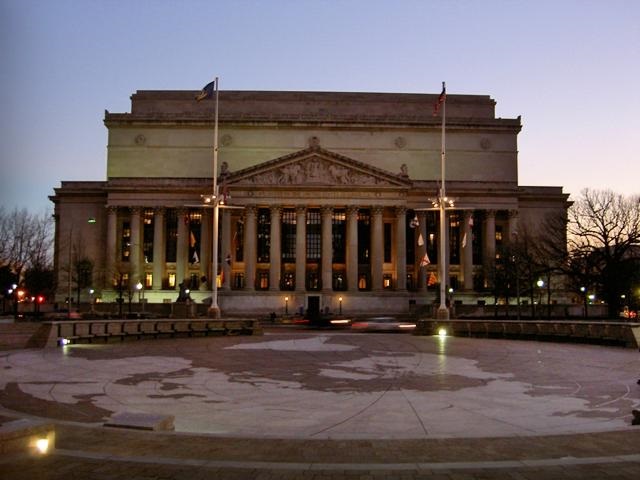}\vspace{2pt}
			\end{minipage}}
	\subfloat[RetinexNet]{
		\begin{minipage}[a]{0.115\textwidth}
			\centering
			\includegraphics[width=0.82in]{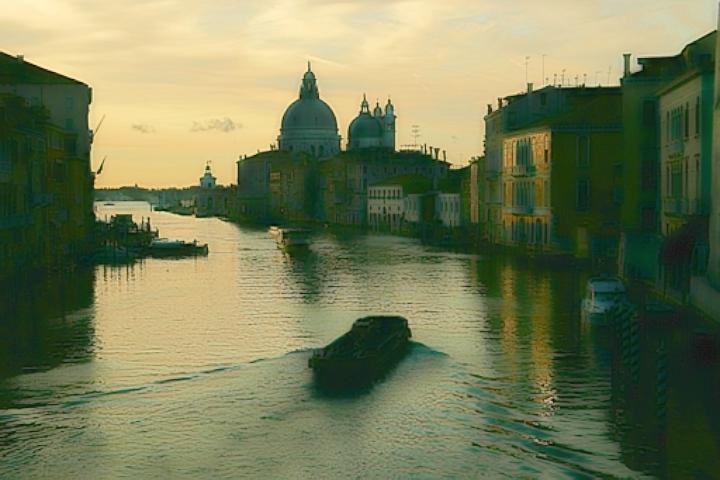}\vspace{2pt}
			\includegraphics[width=0.82in]{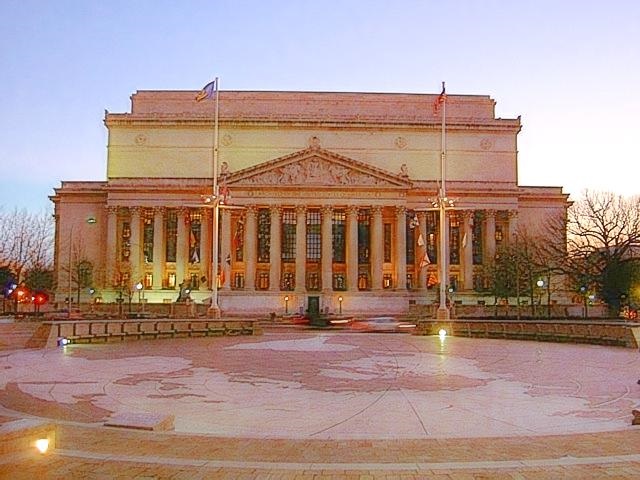}\vspace{2pt}
		\end{minipage}}
	\subfloat[Ours]{
		\begin{minipage}[a]{0.115\textwidth}
			\centering
			\includegraphics[width=0.82in]{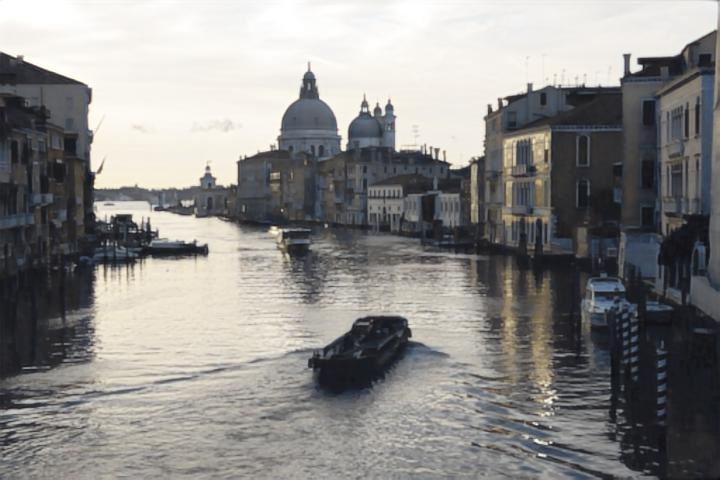}\vspace{2pt}
			\includegraphics[width=0.82in]{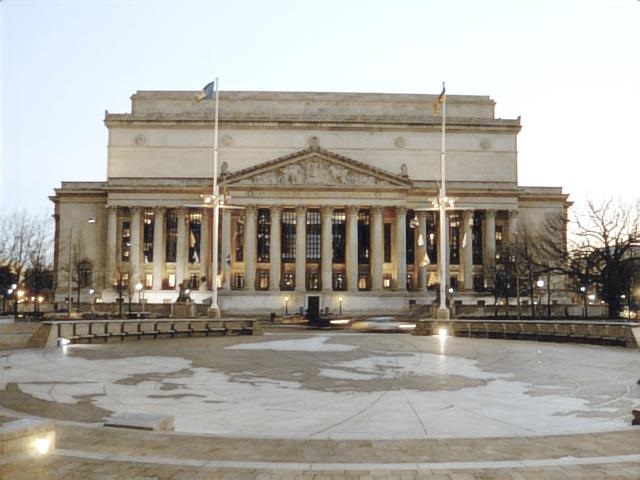}\vspace{2pt}
		\end{minipage}}
	\caption{Examples of low-light image enhanced results. The proposed method can not only improve the contrast of the image but also suppress the noise and artifacts in the dark regions. SRIE generates under-enhancement results and RetinexNet generates results of blur and color distortion.}
	\label{Fig.1}\vspace{-1mm}
\end{figure}

It is noteworthy that many previous methods focused on using the spatial domain information of the image for enhancement, and image processing in frequency domain is also one of the important methods in the image enhancement field. High-frequency information usually represents the image details (\emph{e.g.} contour and edge) or noise, so we proposed a novel Real-low to Real-normal Network for low-light image enhancement, dubbed R2RNet, which utilizes both spatial and frequency information to obtain high visual quality enhanced images. The design of our network is built on the Retinex theory \cite{7}, which includes three subnets: a Decom-Net, a Denoise-Net, and a Relight-Net. The Decom-Net aims to decompose the input low-light image into an illumination map and a reflectance map under the guidance of the Retinex theory. The Denoise-Net takes the decomposition results as input and uses the illumination map as a constraint to suppress the noise in the reflectance to obtain decomposition results with better visual quality. The illumination map obtained by Decom-Net and the reflectance map obtained by Denoise-Net are sent to Relight-Net to improve the contrast and brightness of the image. In Decom-Net and Denoise-Net, we only utilized the spatial information of low-light images, because the purpose of Decom-Net is to decompose the input image into an illumination map and a reflectance map without any further processing. According to the Retinex theory, the reflectance map contains the inherent attributes of the image, so if the frequency information is used to suppress the noise in Denoise-Net, the details of the reflectance map may be suppressed at the same time. Therefore, instead of using the frequency information in Decom-Net and Denoise-Net, we used the spatial information of the image to improve the image contrast and extracted the frequency information of the image based on the fast Fourier transform to better preserve the image details in Relight-Net. By a well-designed network, our method could appropriately enhance the image contrast, preserve more image details, and suppress noise. Moreover, the performance of high-level visual tasks can be effectively improved by using the enhanced results obtained by our method in low-light conditions.

Another difficulty in the low-light image enhancement task is that the learning-based model requires a lot of data for training, and the ability of the model is usually closely related to the quality of the training data. However, it is very difficult to collect sufficient real-world data, especially for paired images. Most learning-based enhancement methods use synthetic datasets for training, which limits their generalization ability in real-world scenarios. As far as we know, the existing real-world low-light image datasets only have the LOL dataset \cite{8} and the SID dataset \cite{9}, but the number of images contained in these two datasets cannot satisfy the training requirement of deep neural networks. Therefore, we collected the first Large-Scale Real-World paired low/normal-light images dataset, named LSRW dataset, for our network training.

The remainder of this paper is organized as follows. Section II briefly reviews the relevant works of low-light enhancement methods, image denoising methods, and low-light image datasets. Section III presents the architecture of the proposed R2RNet and loss function settings. Section IV presents the experimental results, and Section V provides some concluding remarks. 

\section{Related Work}
\subsection{Low-light Image Enhancement methods}
In the past few decades, there have been extensive methods to enhance the contrast of weakly illuminated images. Traditional methods are mainly based on histogram equalization and the Retinex theory. Histogram equalization is a simple but effective image enhancement technology, it takes effect by changing the histogram of the image to enhance the contrast, such as brightness preserving bi-HE \cite{10}. The Retinex theory assumes that the color image observed by human beings can be decomposed into an illumination map and a reflectance map, in which the reflectance map is the inherent attribute of the image and cannot be changed. The purpose of enhancing contrast can be achieved by changing the dynamic range of pixels in the illumination map. MSRCR \cite{11} utilized a multi-scale Gaussian filter to restore color based on the Retinex theory. SRIE \cite{12} proposed a weighted variational model to estimate reflectance and illumination at the same time. MF \cite{13} tried to improve the local contrast of the illumination map and maintain naturalness. BIMEF \cite{14} used a dual-exposure algorithm for image enhancement. In addition to the illumination map and reflectance map, Mading \emph{et al}. \cite{15} added a noise map to form a robust Retinex model for further enhancement and denoising. LIME \cite{16} first estimated the illumination by a prior hypothesis, obtained the estimated illumination through a weighted vibration model, subsequently used BM3D \cite{17} as post-processing. Recently, Liu \emph{et al}. \cite{18} proposed a Retinex-inspired Unrolling with Architecture Search (RUAS) and designed a cooperative reference-free learning strategy to discover low-light prior architectures from a compact search space. Deep learning has been widely used in the field of computer vision and achieves excellent results. Many excellent methods, such as CNN, GAN, \emph{etc}., have made remarkable achievements in a variety of low-level visual tasks, including image de-hazing \cite{19}, \cite{20}, image super-resolution \cite{21}, \cite{22}, \emph{etc}. Many researchers also build learning-based models based on the Retinex theory. MSR-Net \cite{23} utilized different Gaussian convolution kernels to learn the mapping of low/normal-light images. RetinexNet \cite{8} combined the Retinex theory with DeepCNN to estimate and adjust the illumination map to achieve image contrast enhancement and uses BM3D for post-processing to achieve denoising. Zhang \emph{et al}. \cite{24} also designed an effective network based on the Retinex theory to enhance low-light images. Lim \emph{et al}. \cite{25} proposed a deep-stacked Laplacian restorer (DSLR) to recover the global illumination and local details from the original input. Furthermore, some methods that are not based on the Retinex theory are also proposed. Dong \emph{et al}. \cite{21} proposed an algorithm that improves the contrast in dark regions and improves the visual quality by using a de-hazing method. Ying \emph{et al}. \cite{26} used the camera response model for weakly illuminated image enhancement. Lore \emph{et al}. \cite{27} proposed a stacked sparse denoising autoencoder for image contrast enhancement and denoising. Ziaur \emph{et al}. \cite{28} used a bright channel prior to obtain an initial transmission map and adopted L1-norm regularization to refine scene transmission. Guo \emph{et al}. \cite{2} proposed a lightweight network named Zero-DCE, to transform the image enhancement problem into a curve estimation problem. Jiang \emph{et al}. \cite{1} proposed a network based on GAN for low-light image enhancement and used unpaired images for training for the first time. 

The key to Retinex-based methods is the estimation of the illumination map and the reflectance map. Due to the limited decomposition ability, traditional methods often cause over/under-enhancement results. Learning-based methods can get better decomposition results and can properly improve the contrast. It is noteworthy that most learning-based methods only focus on using the spatial information of weakly illuminated images to obtain high-quality normal-light images, and combining spatial and frequency domain information to perform low-light image enhancement can obtain more satisfactory enhanced results. Therefore, our R2RNet uses both spatial and frequency information of the image for enhancement. Spatial information is used for contrast enhancement, and frequency information is used to restore more image details.

\subsection{Denoising methods}
Enhancing weakly illuminated images needs noise suppression in addition to contrast enhancement. The traditional image denoising methods relied on hand-crafted features and used discrete cosine transform or wavelet transform to modify transform coefficients. NLM \cite{29} and BM3D used self-similar patches to achieve outstanding results in image fidelity and visual quality. Image denoising methods based on supervised learning, such as DnCNN-B \cite{30}, FFDNet \cite{31}, and CBDNet \cite{32} utilized the Gaussian mixture model to perform denoising. Mei \emph{et al}. \cite{33} made full use of shallow pixel-level features and self-similarity to achieve a balance between pixel features and semantic features to preserve more details. Kim \emph{et al}. \cite{34} proposed CBAM to focus on learning the difference between noisy images and clear images. Chen \emph{et al}. \cite{35} used GAN to model the noise information extracted from the real noise image and combined the noise blocks generated by the generator with the original clear image to synthesize a new noise image. ADGAN \cite{36} proposed a feature pyramid attention network to improve the ability of network feature extraction when modeling noise.

These methods can achieve impressive denoising results. However, directly using these methods as pre-processing or post-processing of low-light image enhancement methods will result in blurring or noise amplification. To avoid this, our method can perform contrast enhancement and denoising simultaneously.

\subsection{Low-light Image Datasets}
Another difficulty in the low-light image enhancement task is that learning-based models usually require a lot of data, but it is difficult to collect sufficient low-light images. Due to the lack of real-world paired images, most methods use synthetic images based on normal-light images. Lore \emph{et al}. \cite{27} applied gamma correction to each channel to synthesize low-light images. Lv \emph{et al}. \cite{37} used the same image synthesis strategy as LLNet. Lv \emph{et al}. \cite{38} and Wang \emph{et al}. \cite{39} combined linear transformation and gamma transformation to obtain paired images. Wang \emph{et al}. \cite{40} obtained synthetic images by using the camera response function and modeling the noise distribution in the low-light image.

As far as we know, the existing real-world low-light image datasets only have LOL dataset and SID dataset, both of them capture pairs of low/normal-light images by fixing the camera position and changing the ISO and exposure time. LOL dataset contains 500 low/normal-light image pairs. SID dataset contains 5094 short-exposure images and 424 long-exposure images; multiple short-exposure images correspond to one long-exposure image. However, the number of images contained in the above two datasets cannot support the training of DeepCNN, and the SID dataset is mainly suitable for extremely weakly illuminated image enhancement, which is not the same as what we focus on. In order to meet the training requirement of our network, we use a Nikon D7500 camera and a HUAWEI P40 Pro mobile phone to collect real-world paired images to form our LSRW dataset.

\section{LSRW Dataset}
One of the difficulties in the task of low-light image enhancement is the lack of paired low/normal-light images captured in real scenes. The existing real-world paired images datasets are only the LOL dataset and the SID dataset, and SID is mainly suitable for extremely low-light image enhancement, which is not consistent with our concern. To satisfy the training requirements of DeepCNNs and provide support for follow-up researches, we propose the first large-scale real-world paired image dataset, named LSRW dataset. The LSRW dataset contains 5650 paired images captured by a Nikon D7500 camera and a HUAWEI P40 Pro mobile phone. We collected 3170 paired images using the Nikon camera and 2480 paired images using the Huawei mobile phone.

The low-light images can be obtained by reducing ISO and using a shorter exposure time to reduce the amount of light input, while the normal-light images can be obtained by using a larger ISO and a longer exposure time. We chose to collect the LSRW dataset for both indoor and outdoor scenarios. When obtaining low-light images in indoor scenes, the exposure time will be increased to avoid capturing extremely dark images. Similarly, when obtaining normal-light images in outdoor scenes, the exposure time will be reduced to avoid capturing over-exposure images. The ISO value of the low-light condition is fixed to 50, and the normal-light condition is fixed to 100. We can obtain paired low/normal-light images by changing the exposure time. Note that if there are moving objects or camera/phone shaking when reducing the exposure time, the low-light image will be blurred.  Therefore, in order to avoid camera/mobile phone shaking, we use a tripod to fix the position of the camera/mobile phone and adjust the ISO and exposure time through long-range control. At the same time, the scenes we select are static without any moving objects, which can ensure that the captured low-light images will not be blurred. The ISO value of the low-light condition is fixed to 50, and the normal-light condition is fixed to 100. When using Nikon to obtain low-light images, the exposure time is limited to 1/200 to 1/80, while the exposure time of normal-light images is limited to 1/80 to 1/20. When using Huawei to obtain low-light images, the exposure time is limited to 1/400 to 1/100, while the exposure time of normal-light images is limited to 1/100 to 1/15. We selected 5600 paired images from the LSRW dataset for training and the remaining 50 pairs for evaluation. Table I summarizes the LSRW dataset. Fig.2 shows several image pairs in the LSRW dataset, including indoor and outdoor scenes.

\begin{table}[!t]
	\renewcommand{\arraystretch}{1.3}
	\caption{The LSRW Dataset contains 5650 low/normal-light image pairs captured by a Nikon D7500 camera and a Huawei P40 pro mobile phone in real scenes.}
	\label{Table I}
	\centering
	\setlength{\tabcolsep}{2.7mm}{
	\begin{tabular}{cccc}
		\shline
				 & \textbf{Exposure Time(s)} & \textbf{ISO}& \textbf{Image Pairs} \\
		\shline
		Nikon D7500 & {[1/200, 1/20]}  & {[50, 100]}&3170\\
		Huawei P40 Pro & {[1/400, 1/15]}  & {[50, 100]}&2480\\
		\shline
	\end{tabular}}
\end{table}

\begin{figure}[!t]
	\centering
	\subfloat{\includegraphics[width=0.82in]{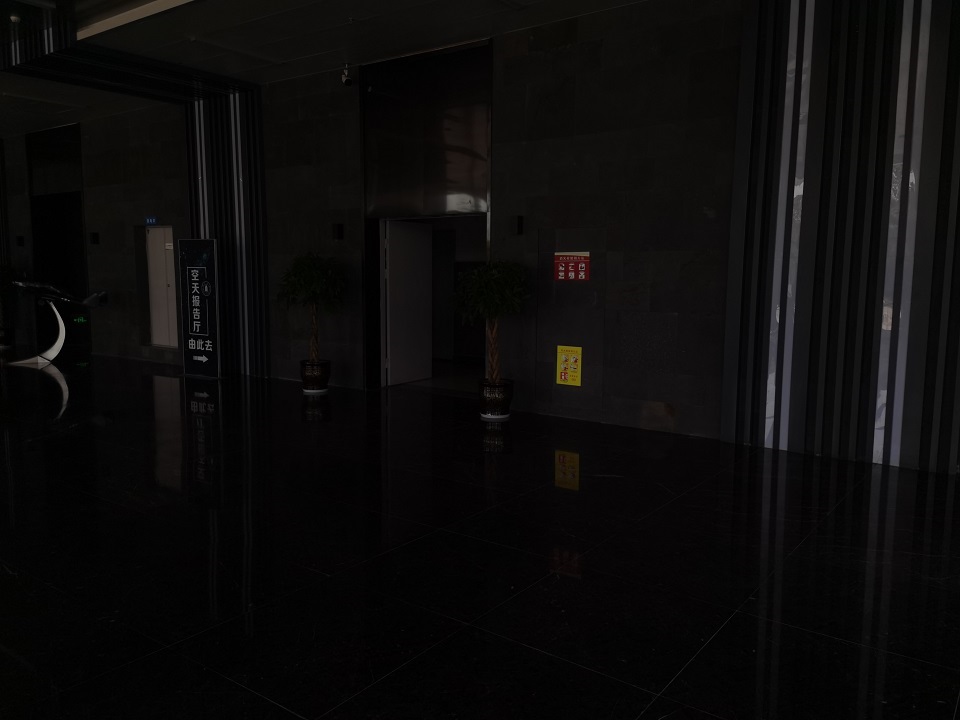}}\
	\subfloat{\includegraphics[width=0.82in]{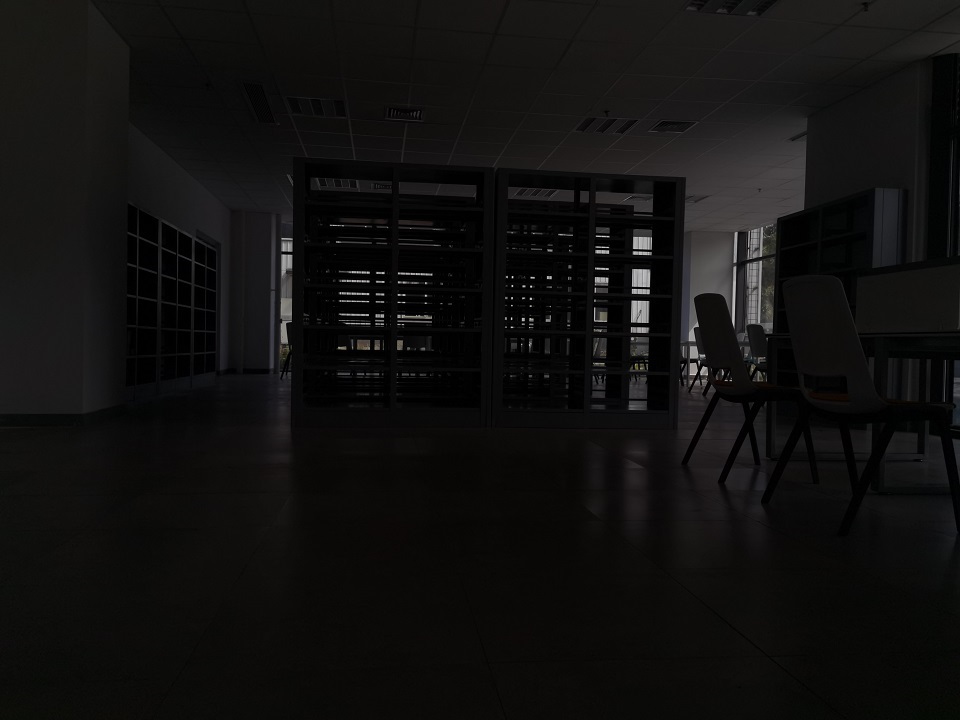}}\
	\subfloat{\includegraphics[width=0.82in]{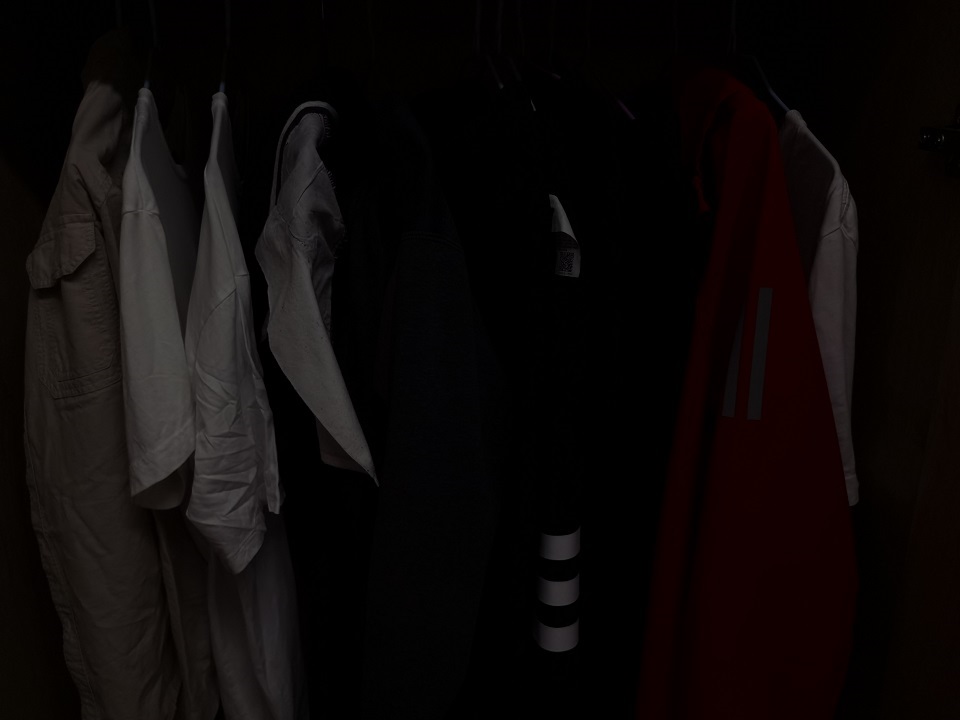}}\
	\subfloat{\includegraphics[width=0.82in]{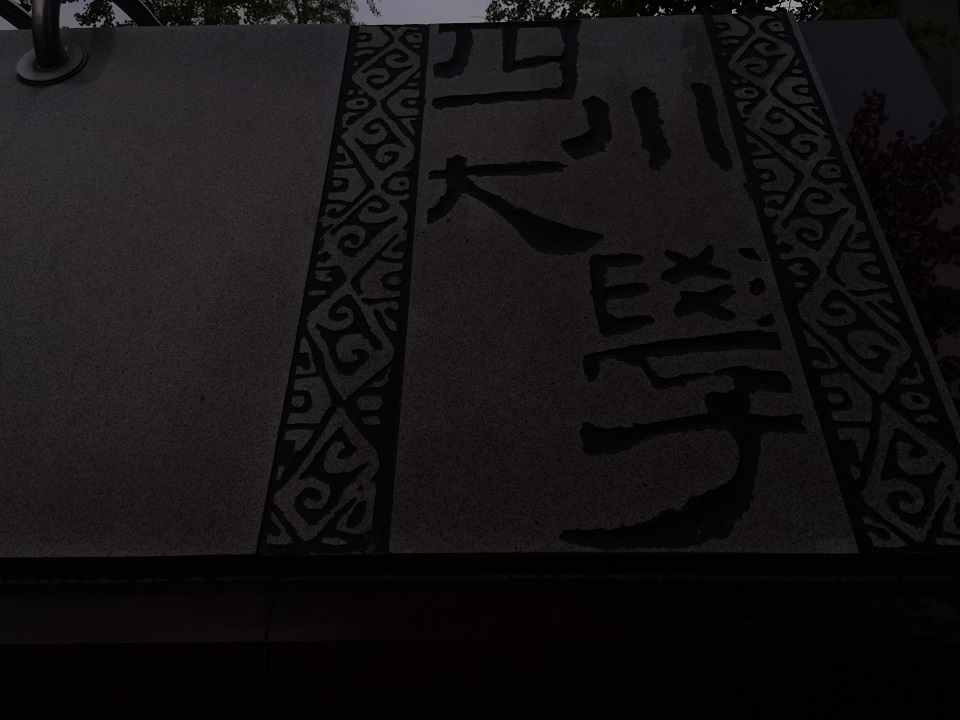}}\vspace{1mm}
	\subfloat{\includegraphics[width=0.82in]{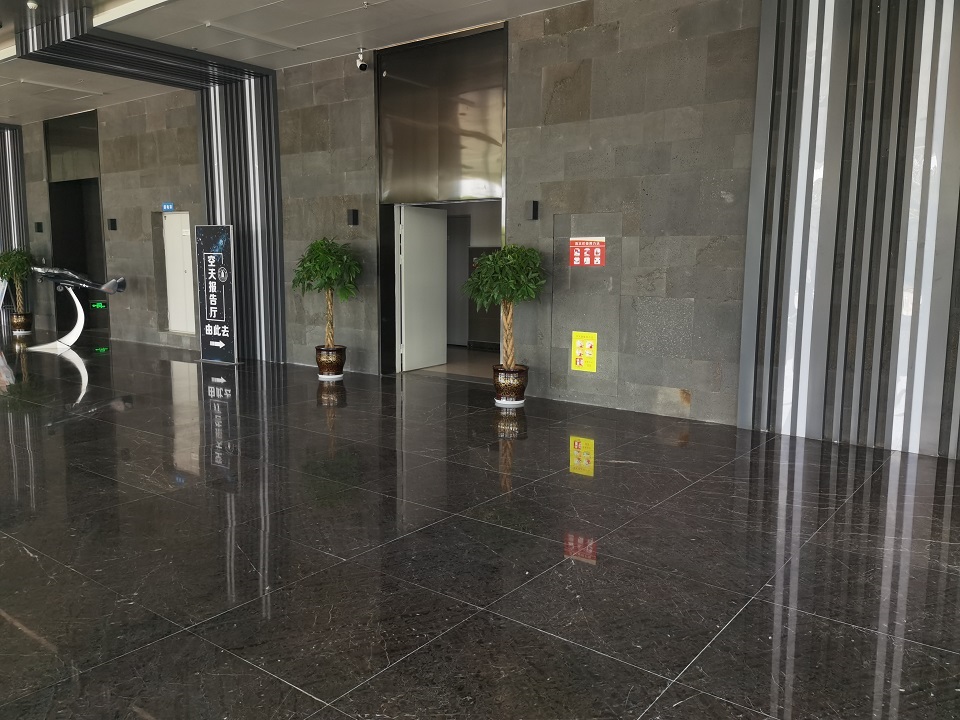}}\
	\subfloat{\includegraphics[width=0.82in]{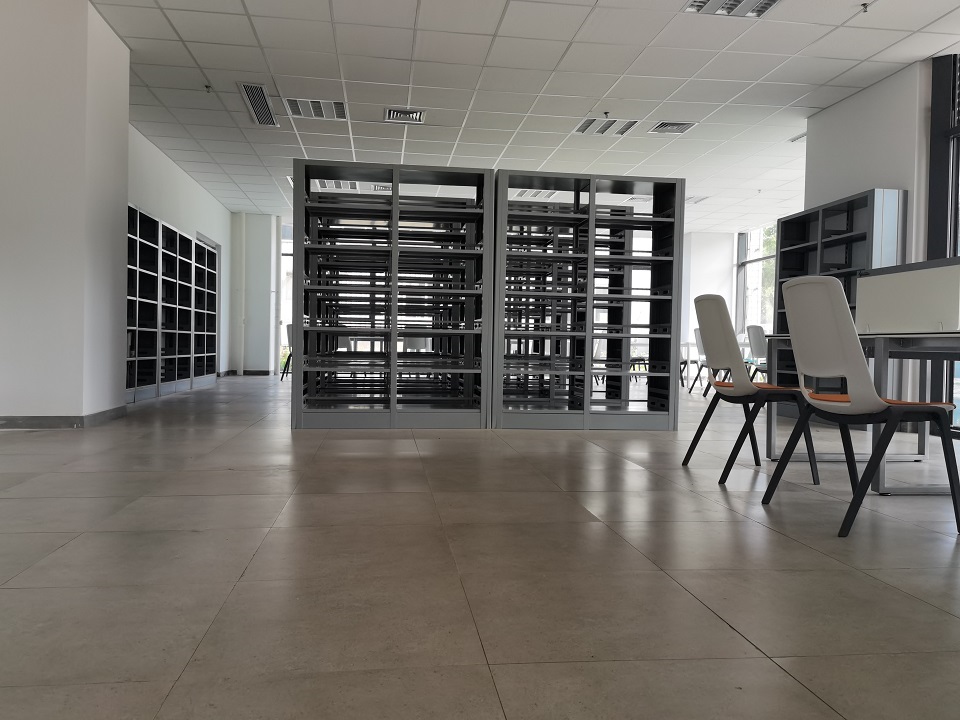}}\
	\subfloat{\includegraphics[width=0.82in]{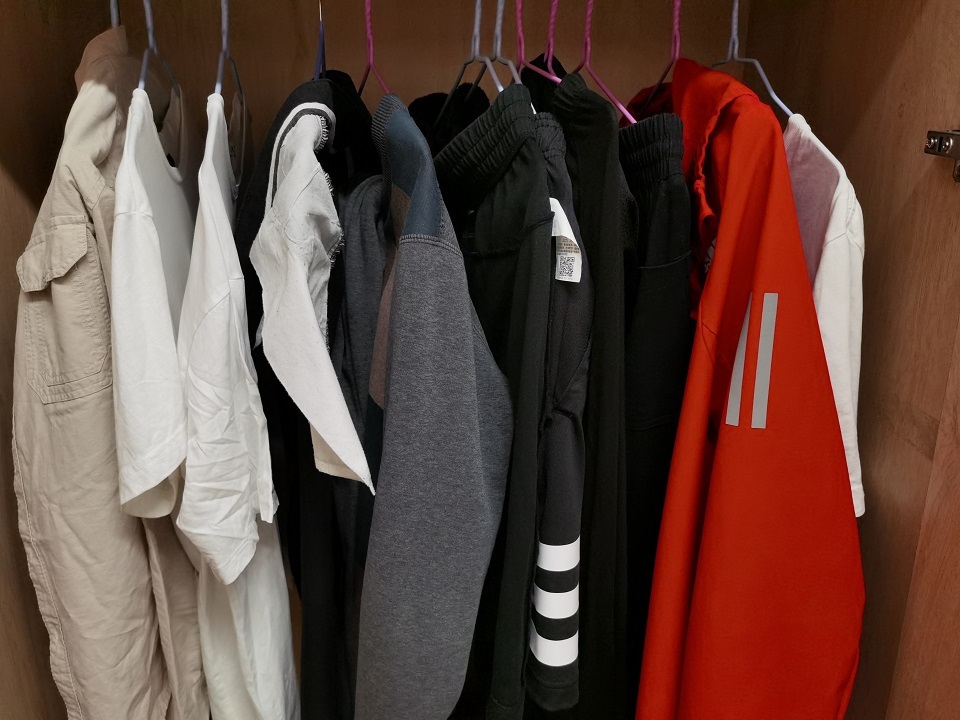}}\
	\subfloat{\includegraphics[width=0.82in]{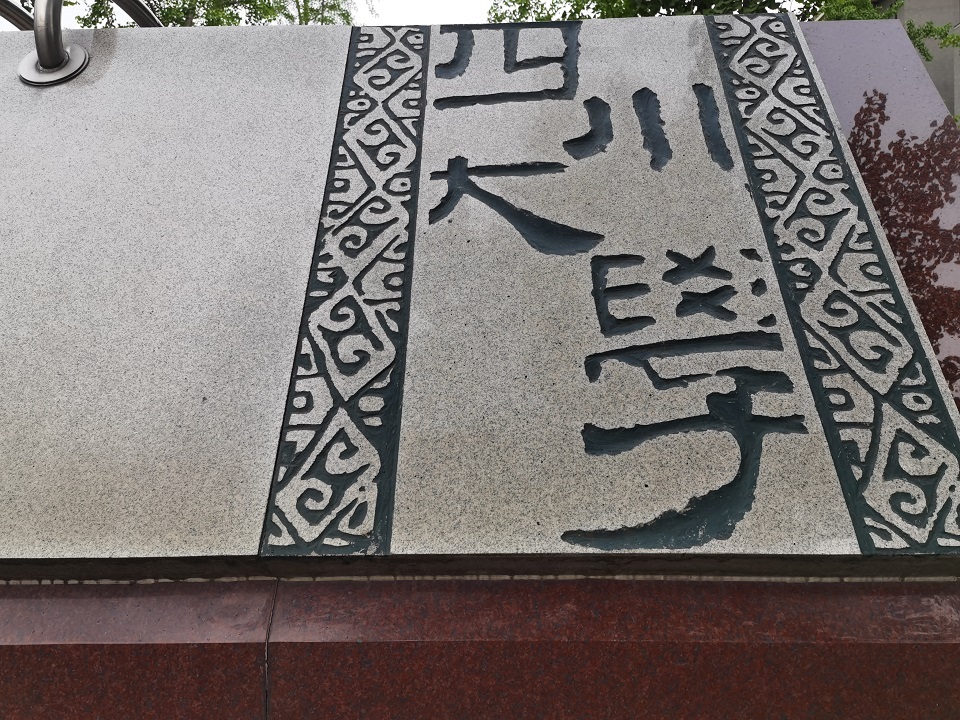}}
	\caption{Several examples for paired low/normal-light images  in the LSRW dataset.}
	\label{Fig.2}\vspace{-1mm}
\end{figure}
\section{Method}
In this section, we will introduce the details of our R2RNet, including the network architecture and the loss function.
\subsection{Network Architecture}
\begin{figure*}
	\centering
	\includegraphics[width=7.2in]{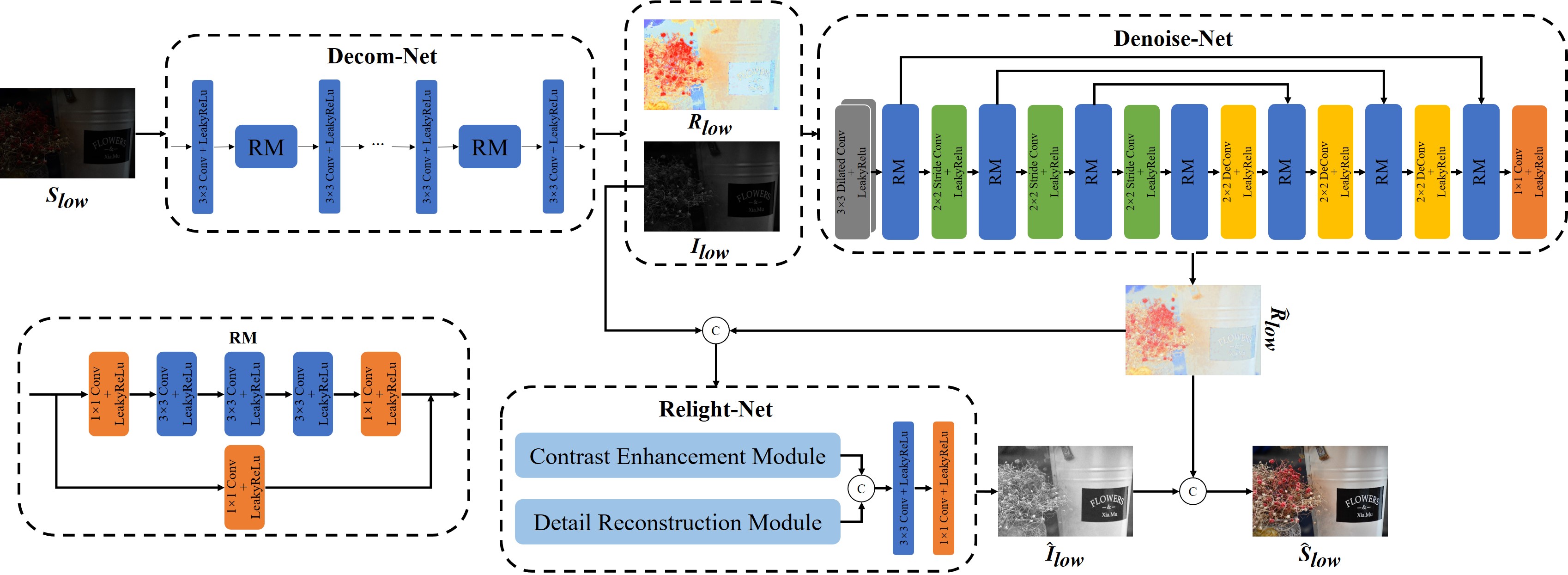}
	\caption{The proposed R2RNet architecture. Our network consists of three subnets: a Decom-Net, a Denoise-Net, and a Enhance-Net, which perform decomposing, denoising, contrast enhancement and detail preservation, respectively. The Decom-Net decomposes the low-light image into an illumination map and a reflectance map based on the Retinex theory. The Denoise-Net aims to suppress the noise in the reflectance map. Subsequently, the illumination map obtained by Decom-Net and the reflectance map obtained by Denoise-Net will be sent to the Relight-Net to improve image contrast and reconstruct details.}
	\label{Fig.3}\vspace{-1mm}
\end{figure*}

We propose a novel deep convolutional neural network, dubbed R2RNet, which consists of three subnets: a Decom-Net, a Denoise-Net, and a Relight-Net. The Decom-Net decomposes the input weakly illuminated image into an illumination map and a reflectance map based on the Retinex theory. The Denoise-Net takes the decomposed results as input and uses the illumination map as a constraint to suppress the noise in the reflectance map. Subsequently, the illumination map obtained by Decom-Net and the reflectance map obtained by Denoise-Net are sent to the Relight-Net to obtain a normal-light image with better visual quality. Therefore, our method can improve the contrast, retain more details, and suppress the noise simultaneously. The network architecture of R2RNet is illustrated in Fig.3. The detailed descriptions are provided below.

\textbf{Decom-Net}: The key of Retinex-based methods is to obtain the high-quality illumination map and reflectance map, the quality of decomposition results will also affect the subsequent enhancement and denoising process. Therefore, it is important to design an efficient network to decompose the weakly illuminated image. Residual network \cite{41} has been widely used in many computer vision tasks and achieved excellent results. Benefiting from the skip connection structure, the residual network can make the deep neural network easier to optimize during the training stage, and not cause gradient disappearance or explosion. Inspired by this, we use multiple Residual Modules (RM) in DecomNet to get better decomposition results. Each RM contains 5 convolutional layers with the kernel size of $\{$1, 3, 3, 3, 1$\}$, the number of kernels is $\{$64, 128, 256, 128, 64$\}$, respectively. And we add a convolutional layer of 64$\times$1$\times$1 at the shortcut connection. There is also a convolution layer of 64$\times$3$\times$3 before and after each RM.

The Decom-Net takes in paired low/normal-light images ($S_{low}$ and $S_{normal}$) each time and learns the decomposition for both low-light and its corresponding normal-light image under the guidance that the low-light image and normal-light image share the same reflectance map. During training, there is no need to provide the ground truth of the reflectance and illumination. Only requisite knowledge including the consistency of reflectance and the smoothness of the illumination map is embedded into the network as loss functions. Note that the illumination map and reflectance map of the normal-light image neither participate in the follow-up training, but only provide a reference for decomposition.

\textbf{Denoise-Net}: Most traditional methods and previous learning-based methods based on the Retinex theory do not take the noise into account after obtaining the decomposition results, it will cause the final enhancement result to be affected by the noise in the reflectance map. Recently, researchers have designed effective models that can suppress noise while enhancing the contrast of low-light images. Inspired by that, we also designed a Denoise-Net to suppress the noise in the reflectance map. Similar to most learning-based methods, our Denoise-Net only uses the spatial information of the image, because eliminating noise by suppressing high-frequency signals in the reflectance map may result in the loss of inherent details.

U-Net \cite{42} has achieved excellent results in a large number of computer vision tasks due to its excellent structural design. In the field of low-light image enhancement, a large number of networks have adopted the U-Net as the main architecture or a part of it. Chen \emph{et al}. \cite{9} directly uses U-Net to enhance the image without any modification to the network and achieve excellent results. Inspired by the residual network, Res-UNet \cite{43} substitutes a module with residual connections for each sub-module of U-Net. However, U-Net and Res-UNet use multiple max-pooling layers in the feature extraction stage, and the max-pooling layer will lead to the loss of feature information, this is what we do not want. Inspired by \cite{44}, we replace the max-pooling layers with stride convolutional layers, which will slightly increase the network parameters, but improve the performance. Both U-Net and Res-UNet belong to "shallow-wide" architecture, Li \emph{et al}. \cite{45} demonstrate that "deep-narrow" architecture is more efficient, so we replace each sub-module of UNet with RM to build "deep-narrow" Res-UNet, which is named DN-ResUnet in this paper. The RM used in Denoise-Net is slightly different from that in Decom-Net, the number of convolutions is maintained at 128 without increasing, except for the last 1$\times$1 convolutional layer. Moreover, we use dilated convolution in the first two layers of the network to extract more feature information. As shown in Fig.5, the illumination map obtained by our Denoise-Net retains the original image details while suppressing noise.

\textbf{Relight-Net}: After getting the decomposition results, it is necessary to improve the contrast of the illumination map to obtain a high visual quality result, which is the purpose of Relight-Net design. Inspired by the effectiveness of combing spatial and frequency information to restore high-quality sharp images in other image restoration tasks \cite{46}, our Relight-Net consists of two modules: Contrast Enhancement Module (CEM) and Detail Reconstruction Module (DRM). The CEM uses spatial information for contrast enhancement, its architecture is similar to Denoise-Net, we also utilize multi-scale fusion, concatenate the output of each deconvolutional layer in the expansive path to reduce the loss of feature information. The DRM extracts frequency information based on the Fourier transform to recover more details. The Fourier transform aims to obtain the distribution of signals in the frequency domain. Digital image is also a kind of signal, the Fourier transform can transform the image from spatial domain to frequency domain, and inverse Fourier transform can transform the image from frequency domain to spatial domain. Therefore, we can get the spectrum information of the image through Fourier transform. The high-frequency signals represent intense change contents in the image, \emph{i.e.} details or noise, and the low-frequency signals represent smooth change contents that do not change frequently, \emph{i.e.} background. We can restore more details by enhancing the high-frequency signal in the image, so as to recover the sharp image from the degraded image. 

\begin{figure*}
	\centering
	\includegraphics[width=7.1in]{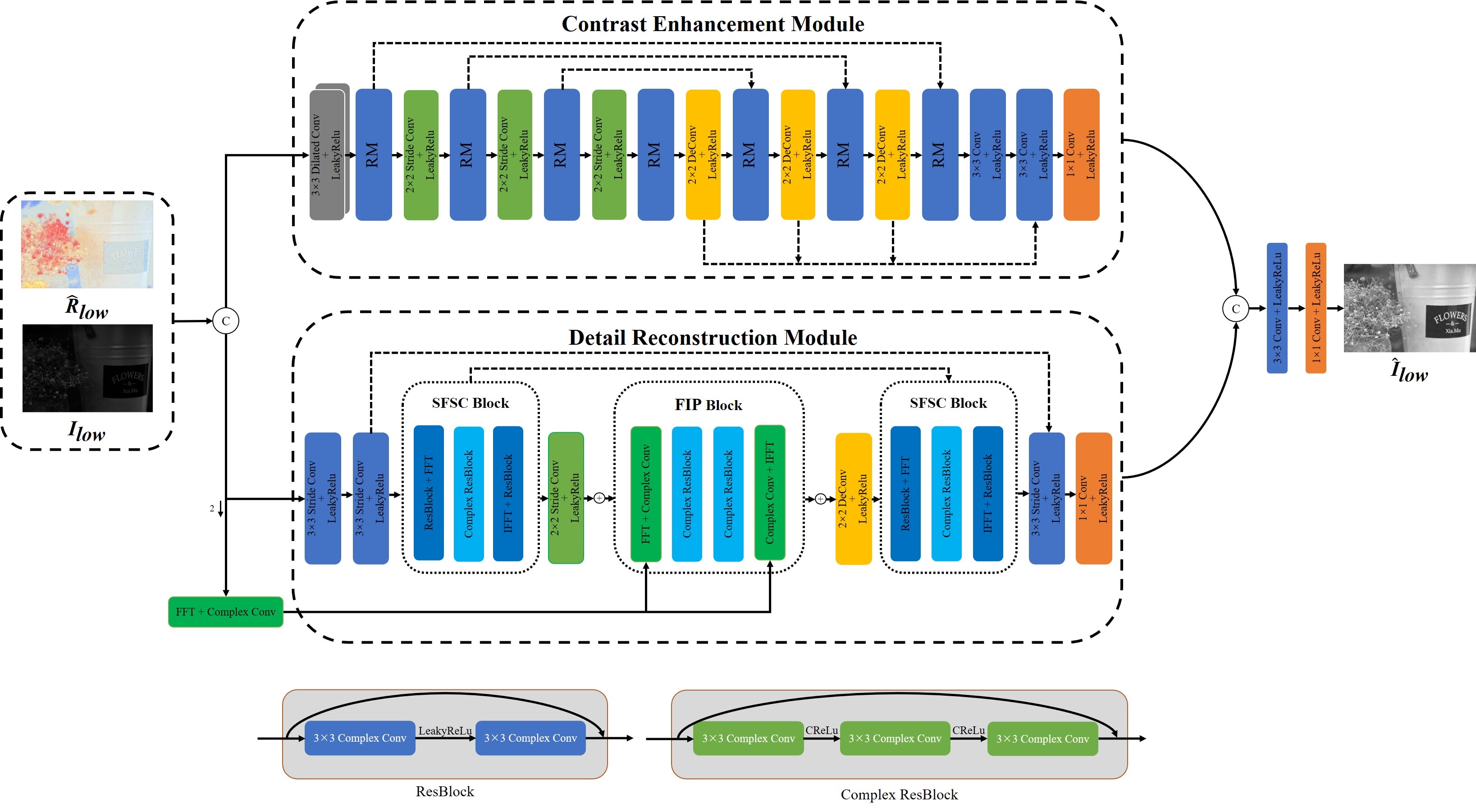}
	\caption{The proposed Relight-Net architecture. The Relight-Net consists of two modules: Contrast Enhancement Module (CEM) and Detail Reconstruction Module (DRM). CEM uses spatial information for contrast enhancement and DRM uses frequency information to preserve image details.}
	\label{Fig.4}\vspace{-1mm}
\end{figure*}

The original Fourier transform is time-consuming, so we use the fast Fourier transform in this paper. After Fourier transform, we will get a matrix with the same size as the original image. The points in the matrix describe the frequency domain information of the image. Each point is a complex number $A + jB$, its modulus $\sqrt{A_2 + B_2}$ describes amplitude, and its direction $\arctan\frac{B}{A}$ describes phase angle. If we want to utilize the frequency domain information of the image to realize detail reconstruction, we need to process the obtained complex matrix. Chiheb \emph{et.al} \cite{47} proposed the key atomic components, complex convolution, complex batch normalization, and complex-valued activations, to form complex-valued deep neural networks and achieve state-of-the-art performance on many computer vision tasks and audio-related tasks. The complex convolution convolves a complex filter matrix $W = A + jB$ by a complex-valued vector $h=x + jy$ where A and B are real-valued matrices and x and y are real-valued vectors. After convolving the vector h by the filter W we can obtain $W*h =(A*x-B*y)+i(B*x+A*y)$. And the complex Relu (CRelu) uses separate ReLUs on both of the real and the imaginary part of a neuron, \emph{i.e.} $CReLU(W)=ReLU(A)+iReLU(B)$. Therefore, we chose the complex convolution and the CRelu to form our DRM, so that we can aggerate the amplitude and phase information in the frequency domain. 

Our DRM consists of two Spatial-Frequency-Spatial Conversion Blocks (SFSC block) and one Frequency Information Processing Block (FIP block). The SFSC block aims to aggerate the frequency and spatial domain information flow. The SFSC block first processes the features in the spatial domain by using the first Resblock and the output feature is converted into the frequency domain by using the fast Fourier transform. Subsequently, the complexResblock is used to process the frequency domain information and finally uses the inverse Fourier transform to transform the frequency domain information into the spatial domain, which can maximize the exchange of information in the spatial domain and the frequency domain. The FIP block is used to simulate the high-pass filter to enhance the image edge contour to realize detail reconstruction. The input of FIP block contains feature-level and image-level frequency signals to reduce the information loss caused by the conversion between spatial and frequency domain information. The feature-level signal denotes the output of SFSC block and the image-level signal can be obtained by directly mapping the input image to the frequency domain based on the fast Fourier transform. The outputs of CEM and DRM will be combined as the enhanced illumination map. Note that the number of output channels of DRM and CEM is 64, so we add a 3$\times$3 convolution layer and a 1$\times$1 convolution layer for dimension reduction. The architecture of Relight-Net is illustrated in Fig.4.

The input of Relight-Net is the illumination map ($I_{low}$) obtained by Decom-Net and the reflectance map ($\hat{R}_{low}$) obtained by Denoise-Net, and the output is the enhanced illumination map ($\hat{I}_{low}$). Finally, the denoised reflectance and the enhanced illumination are combined by element-by-element multiplication as the final result, which can be described as:
\begin{equation}
	\hat{S}_{low} = \hat{I}_{low} \circ \hat{R}_{low}
\end{equation}

The decomposition results obtained by our method are illustrated in Fig.5. The reflectance map obtained by Denoise-Net retains the original image details while suppressing the noise, and the Relight-Net properly improves the contrast of the illumination map and retains more details. 

\begin{figure*}[!h]
	\centering
	\subfloat[Input]{\includegraphics[width=2.3in]{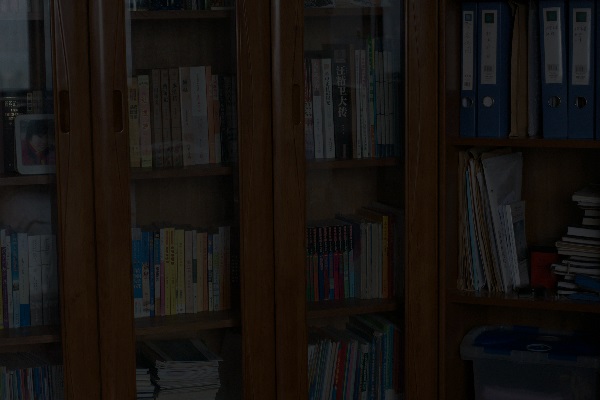}}\
	\subfloat[Illumination Map]{\includegraphics[width=2.3in]{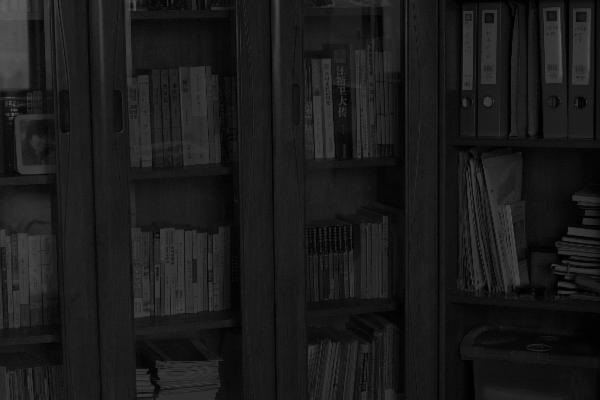}}\
	\subfloat[Reflectance Map]{\includegraphics[width=2.3in]{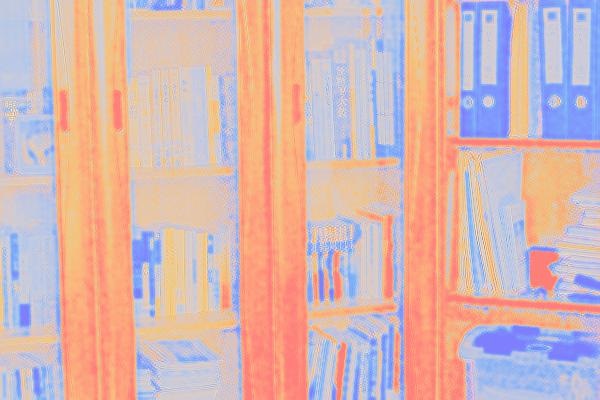}} \\ \vspace{-1mm}
	\subfloat[Enhanced Result]{\includegraphics[width=2.3in]{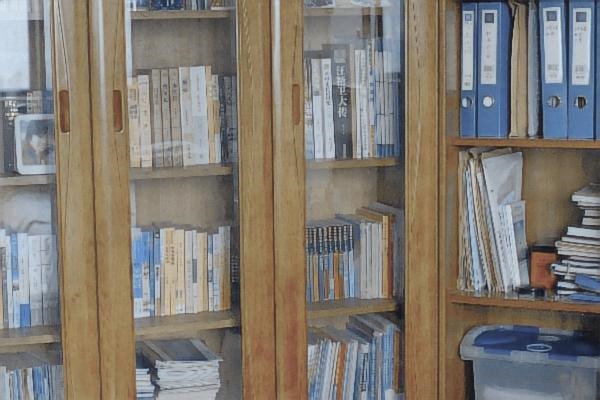}}\
	\subfloat[Enhanced Illumination Map]{\includegraphics[width=2.3in]{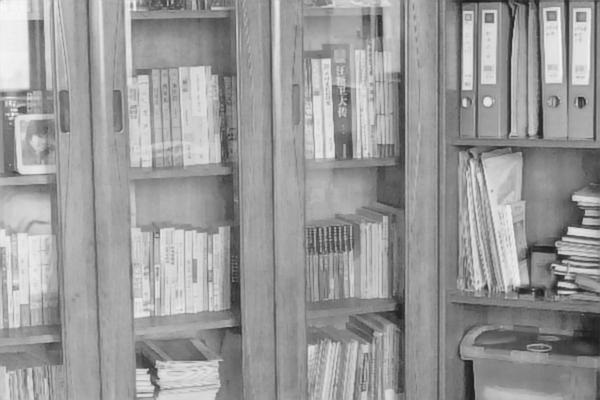}}\
	\subfloat[Denoised Reflectance Map]{\includegraphics[width=2.3in]{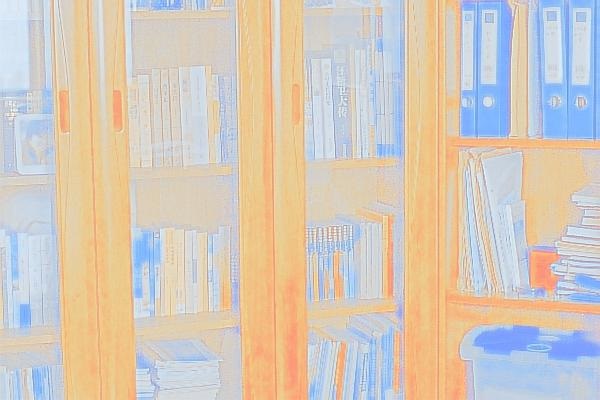}}\
	\caption{The decomposition results obtained by our method. The illumination map and reflectance map are obtained by the Decom-Net, the enhanced illumination map is obtained by Relight-Net, and the denoised reflectance map is obtained by Denoise-Net.}
	\label{Fig.5}\vspace{-1mm}
\end{figure*}

\subsection{Loss Function}
In the training phase, the three subnets are trained separately, so the whole loss function consists of three parts: the decomposition loss $\mathcal{L}_{Dc}$ the denoise loss $\mathcal{L}_{Dn}$ , and the relight loss $\mathcal{L}_{Re}$ . Each of them consists of two parts: content loss and perceptual loss \cite{48}.

\textbf{Decomposition loss}: Our decomposition loss contains two parts: content loss $\mathcal{L}_{Dc-con}$, and perceptual loss $\mathcal{L}_{Dc-per}$ . We use the L1 loss as content loss,
\begin{equation}
	\begin{aligned}
	\mathcal{L}_{Dc-con}=\sum\limits_{i = 1}^N {{\rm{|}}}R_{low} \circ I_{low}-S_{low}{\rm{|}} + \\ \sum\limits_{i = 1}^N {{\rm{|}}}R_{nor} \circ I_{nor}-S_{nor}{\rm{|}} + \\
	\lambda_{1}\sum\limits_{i = 1}^N {{\rm{|}}}R_{nor} \circ I_{low}-S_{low}{\rm{|}} + \\ \lambda_{2}\sum\limits_{i = 1}^N {{\rm{|}}}R_{low} \circ I_{nor}-S_{nor}{\rm{|}} 
	\end{aligned}
\end{equation}
and we calculate the perceptual loss based on features extracted from a VGG-16 pre-trained model, and in contrast to previous methods, we adopt features before rather than after the activation layer,
\begin{equation}
	\begin{aligned}
		\mathcal{L}_{Dc-per}=\frac{1}{C_{j}H_{j}W_{j}}{\rm{||}}\phi_j (R_{low} \circ I_{low})-\phi_j (S_{low}){\rm{||}}^{2}_{2} + \\ 
		\frac{1}{C_{j}H_{j}W_{j}}{\rm{||}}\phi_j (R_{nor} \circ I_{nor})-\phi_j (S_{nor}){\rm{||}}^{2}_{2}
	\end{aligned}
\end{equation}

The decomposition loss is formulated as:
\begin{equation}
	\mathcal{L}_{Dc} = \mathcal{L}_{Dc-con} + \lambda_{3} \mathcal{L}_{Dc-per}
\end{equation}

\textbf{Denoise loss}: Similar to the decomposition loss, denoise loss contains two parts: content loss $\mathcal{L}_{Dn-con}$, and perceptual loss $\mathcal{L}_{Dn-per}$. We also adopt L1 loss as content loss,
\begin{equation}
	\begin{aligned}
		\mathcal{L}_{Dn-con}=\sum\limits_{i = 1}^N {{\rm{|}}}\hat{R}_{low} -R_{nor}{\rm{|}} 
	\end{aligned}
\end{equation}
and use features before the activation layer extracted from a VGG-16 pre-trained model to calculate the perceptual loss.

\begin{equation}
	\begin{aligned}
		\mathcal{L}_{Dn-per}=\frac{1}{C_{j}H_{j}W_{j}}{\rm{||}}\phi_j (\hat{R}_{low})-\phi_j (R_{nor}){\rm{||}}^{2}_{2} 
	\end{aligned}
\end{equation}

The denoise loss is formulated as:
\begin{equation}
	\mathcal{L}_{Dn} = \mathcal{L}_{Dn-con} + \lambda_{4} \mathcal{L}_{Dn-per}
\end{equation}

\textbf{Relight Loss}: Relight loss contains content loss, perceptual loss, and detail preserve loss. We use the same strategy as decomposition loss and denoise loss to build content loss and perceptual loss.
\begin{equation}
	\begin{aligned}
		\mathcal{L}_{Re-con}=\sum\limits_{i = 1}^N {{\rm{|}}}\hat{S}_{low} -S_{nor}{\rm{|}} 
	\end{aligned}
\end{equation}
\begin{equation}
	\begin{aligned}
		\mathcal{L}_{Re-per}=\frac{1}{C_{j}H_{j}W_{j}}{\rm{||}}\phi_j (\hat{S}_{low})-\phi_j (S_{nor}){\rm{||}}^{2}_{2} 
	\end{aligned}
\end{equation}

Moreover, since we used the frequency information in the Relight-Net, we proposed a novel frequency loss to help Relight-Net recover more details. The enhanced image and the sharp image are converted into frequency domain by fast Fourier transform, and the Wasserstein distance is used to minimize the difference between the real part and imaginary part of enhanced result and the ground truth in frequency domain. The frequency loss is formulated as:
\begin{equation}
	\begin{aligned}
		\mathcal{L}_{Re-fre}= \frac{1}{N^2} \sum\limits_{i = real}^{imag} inf_{\gamma \sim \prod (\hat{S}_{low}^{i}, S_{nor}^{i})} \mathbb{E}_{(x, y)\sim [||\hat{S}_{low}^{i}- S_{nor}^{i}||]}
	\end{aligned}
\end{equation}

The relight loss is formulated as:
\begin{equation}
	\mathcal{L}_{Re} = \mathcal{L}_{Ren-con} + \lambda_{5} \mathcal{L}_{Re-per}+ \lambda_{6} \mathcal{L}_{Re-fre}
\end{equation}

\section{Experiments}
\subsection{Implementation Details}
Our implementation is done with PyTorch. The proposed network can be quickly converged after being trained for 20 epochs on a 1080Ti GPU with our LSRW dataset. We use the Adam \cite{49} optimizer with $lr=10^{-3}$, $\beta_1=0.9$, and $\beta_2=0.999$. The batch size and patch size are set to 4 and 96, respectively. We also use the learning rate decay strategy, which reduces the learning rate to $10^{-4}$ after 10 epochs. $\lambda_{1}$ and $\lambda_{2}$ in Eq.2 are set to 0.01, $\lambda_{3}$ in Eq.4, $\lambda_{4}$ in Eq.7, and $\lambda_{5}$ in Eq.11 are set to 0.1, $\lambda_{6}$ in Eq.11 is set to 0.01. For more implementation details of our network, please refer to the code we are going to release.

\begin{figure*}[!h]
	\centering
	\subfloat[Input]{\includegraphics[width=1.75in]{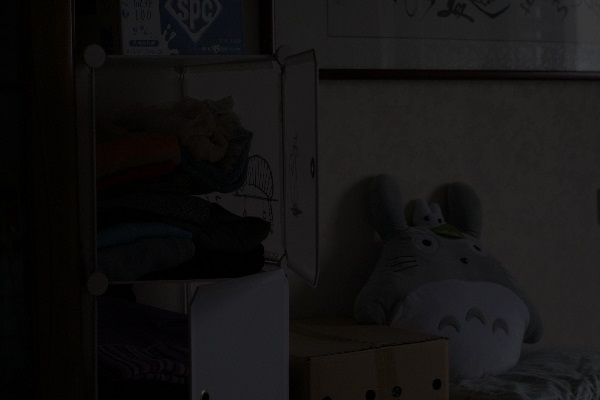}}\
	\subfloat[Dong]{\includegraphics[width=1.75in]{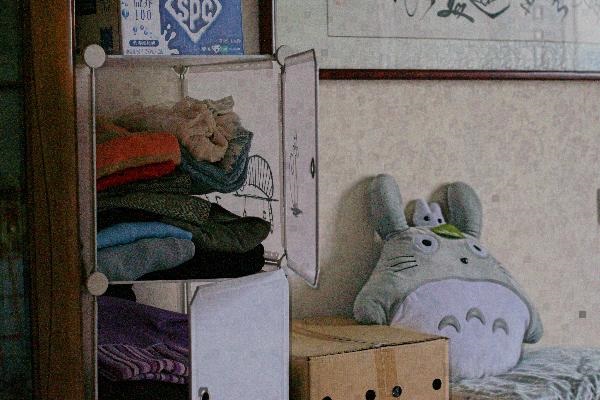}}\
	\subfloat[SRIE]{\includegraphics[width=1.75in]{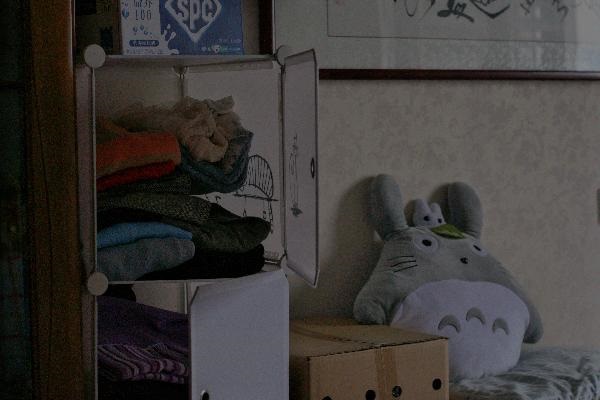}}\
	\subfloat[NPE]{\includegraphics[width=1.75in]{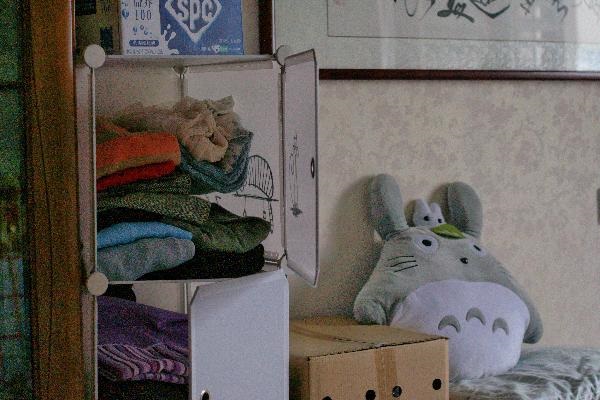}}\\ \vspace{-3mm}
	\subfloat[RetinexNet]{\includegraphics[width=1.75in]{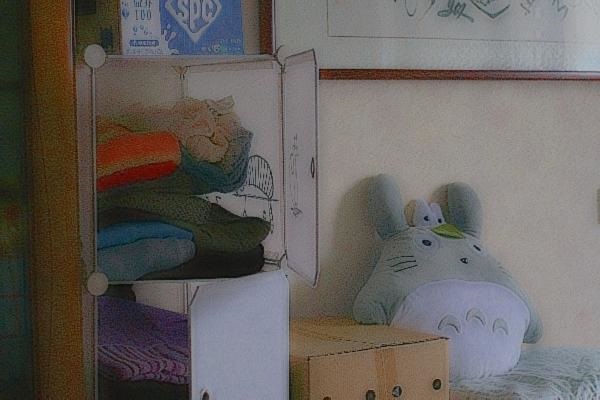}}\
	\subfloat[MBLLEN]{\includegraphics[width=1.75in]{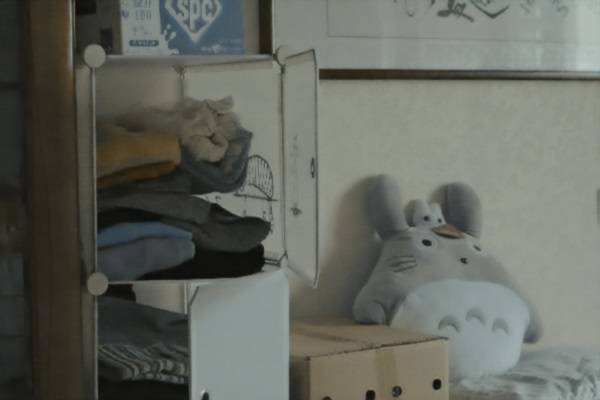}}\
	\subfloat[EG]{\includegraphics[width=1.75in]{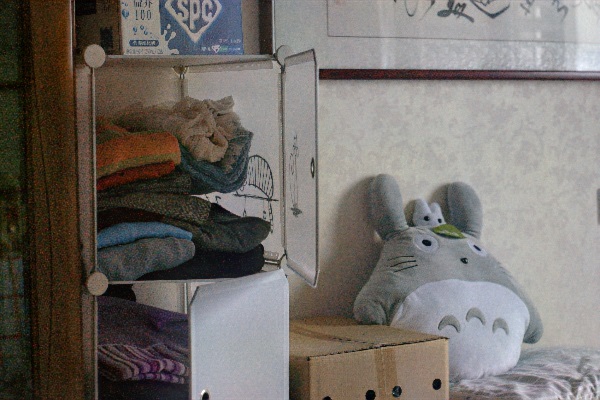}}\
	\subfloat[Ours]{\includegraphics[width=1.75in]{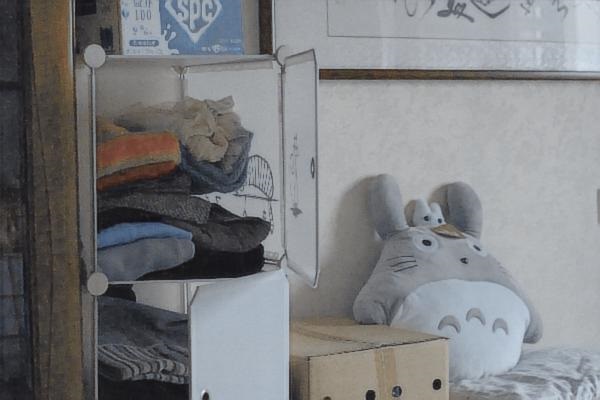}}\vspace{-4mm}
\end{figure*}
\begin{figure*}[!h]
	\centering
	\subfloat[Input]{\includegraphics[width=1.75in]{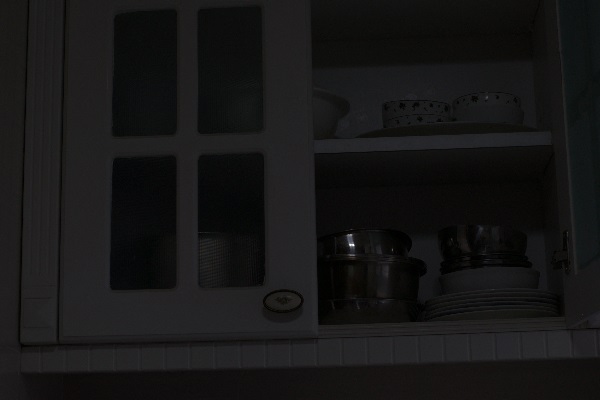}}\
	\subfloat[Dong]{\includegraphics[width=1.75in]{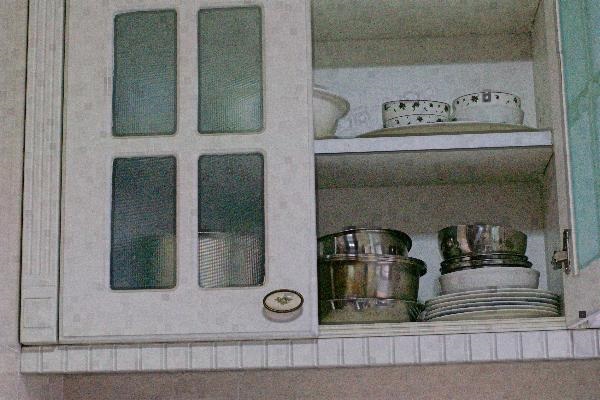}}\
	\subfloat[BIMEF]{\includegraphics[width=1.75in]{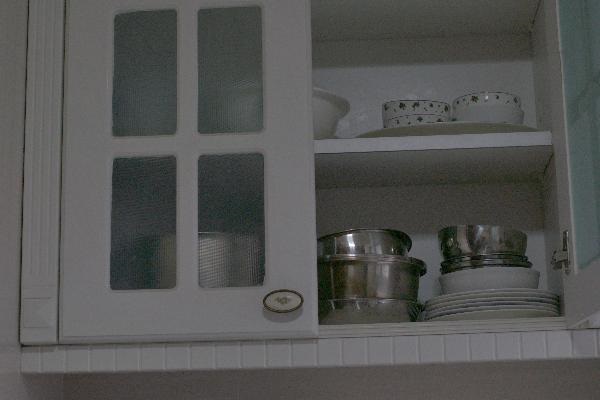}}\
	\subfloat[MSRCR]{\includegraphics[width=1.75in]{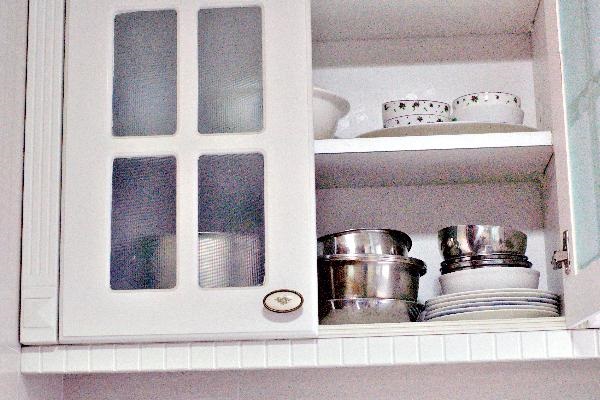}}\\ \vspace{-3mm}
	\subfloat[LIME]{\includegraphics[width=1.75in]{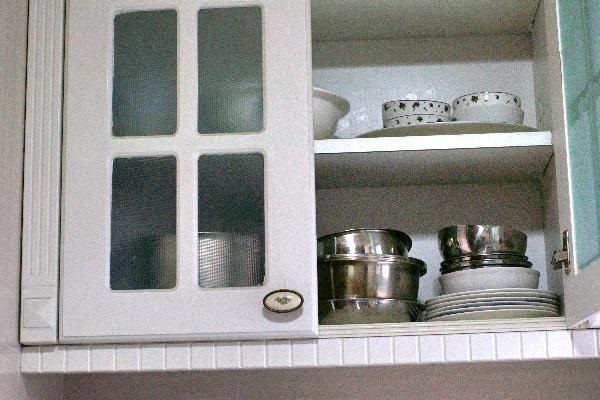}}\
	\subfloat[EG]{\includegraphics[width=1.75in]{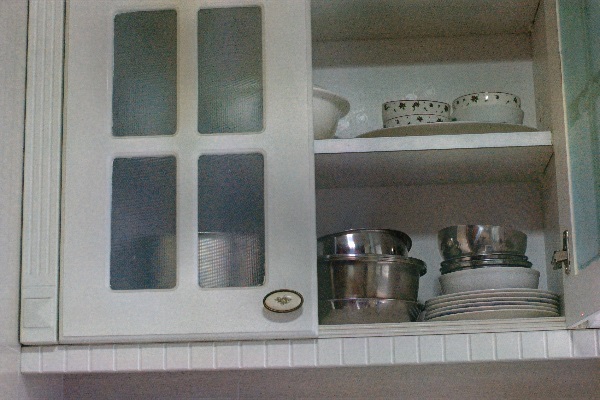}}\
	\subfloat[Zero-DCE]{\includegraphics[width=1.75in]{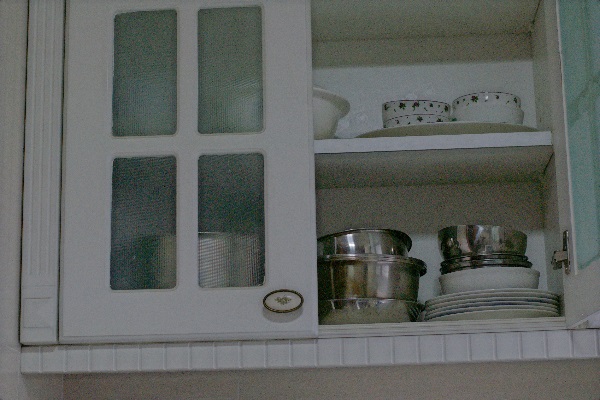}}\
	\subfloat[Ours]{\includegraphics[width=1.75in]{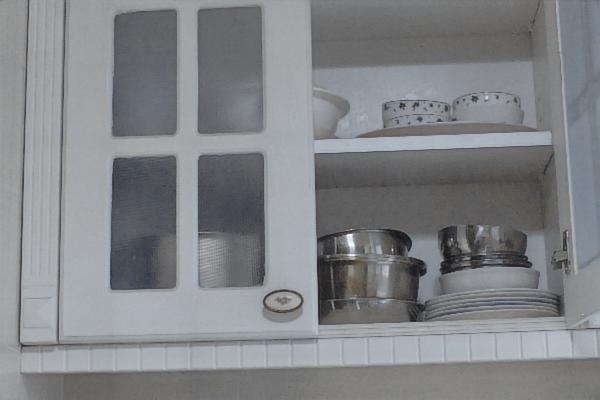}}\vspace{-1mm}
		\caption{Visual Comparison with state-of-the-art low-light image enhancement methods on the LOL dataset. Please zoom in for a better review. EG denotes EnlightenGAN.}
	\label{Fig.6}\vspace{-1mm}
\end{figure*}

\subsection{Comparison with State-of-the-Arts on the Real Datasets}

We compare the proposed method with the existing state-of-art methods (MF, Dong, NPE, SRIE, BIMEF, MSRCR, LIME, RetinexNet, DSLR, MBLLEN, EnlightenGAN, and Zero-DCE) on six publicly available datasets, including LOL, LIME, DICM \cite{50}, NPE \cite{51}, MEF \cite{52}, and VV\footnote{https://sites.google.com/site/vonikakis/datasets}. For fair comparison, we use the released codes of these methods without any modification, and use our LSRW dataset to train supervised learning-based methods, including Retinexnet, MBLLEN, and DSLR. Since Zero-DCE and EnlightenGAN use unpaired data for training, we use their published pre-trained model for comparison. The LOL dataset captured 500 pairs of real low/normal-light images by changing the camera’s exposure time and ISO. This is the only existing real low/normal-light image dataset for low-light image enhancement (the SID dataset is used for extremely low-light image enhancement). The results are shown in Table II. It can be seen that our method outperforms the existing state-of-the-art methods on the LOL dataset for both PSNR and SSIM, the proposed R2RNet achieves the best performance with an average PSNR score of 20.207dB and SSIM score of 0.816, which exceed the second-best approach (MBLLEN) by 1.347dB (20.207-18.860) on PSNR and 0.062 (0.816-0.754) on SSIM. The visual comparison is illustrated in Fig.6. It can be seen  that some traditional methods (such as SRIE, NPE) will cause under-enhancement results, while other methods based on the Retinex theory (such as LIME, RetinexNet) will blur the details or amplify the noise. The enhancement results generated by our method can not only improve the local and global contrast, have clearer details, but also suppress the noise well, which demonstrates that our method can enhance the image contrast and suppress noise simultaneously. Please zoom in to compare more details.
\begin{figure*}[!h]
	\centering
	\subfloat[Input]{\includegraphics[width=1.75in]{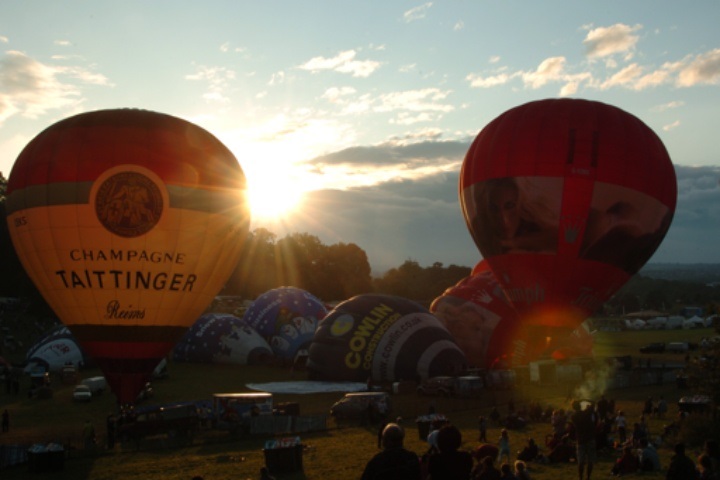}}\
	\subfloat[BIMEF]{\includegraphics[width=1.75in]{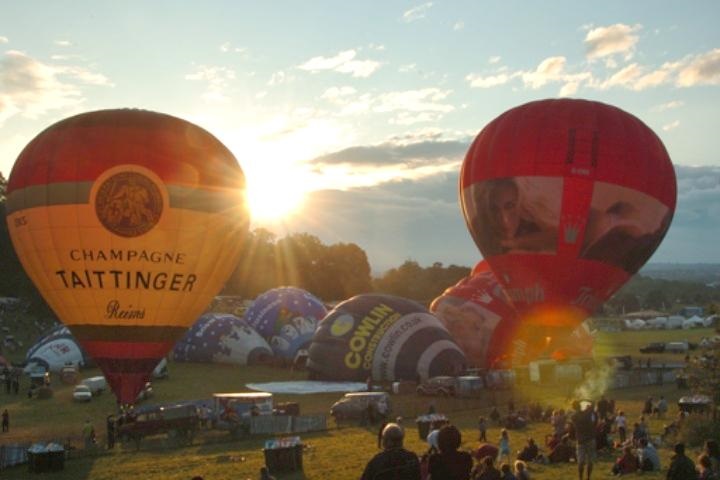}}\
	\subfloat[SRIE]{\includegraphics[width=1.75in]{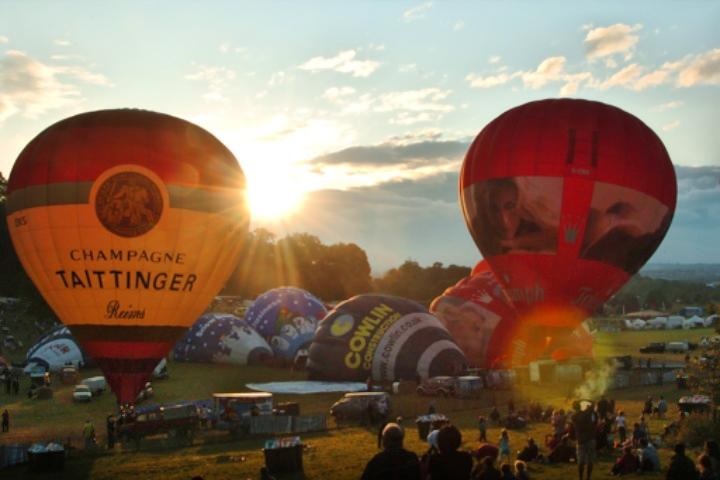}}\
	\subfloat[LIME]{\includegraphics[width=1.75in]{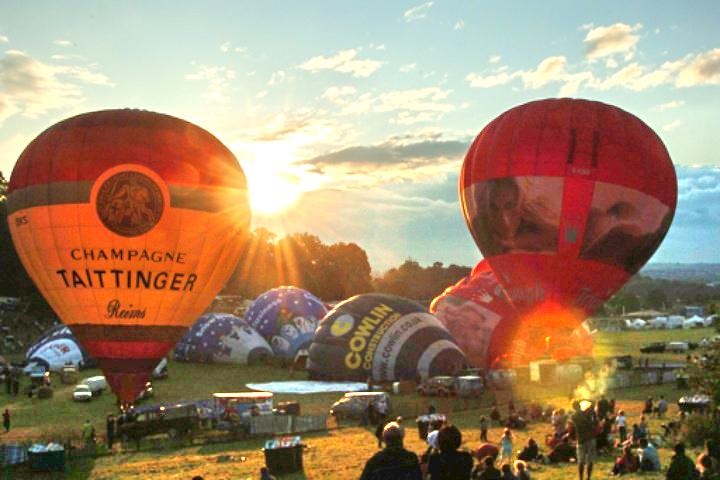}}\\ \vspace{-3mm}
	\subfloat[RetinexNet]{\includegraphics[width=1.75in]{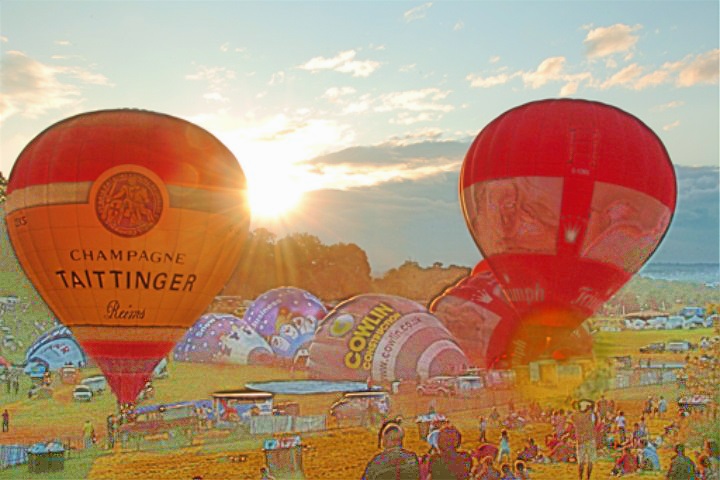}}\
	\subfloat[MBLLEN]{\includegraphics[width=1.75in]{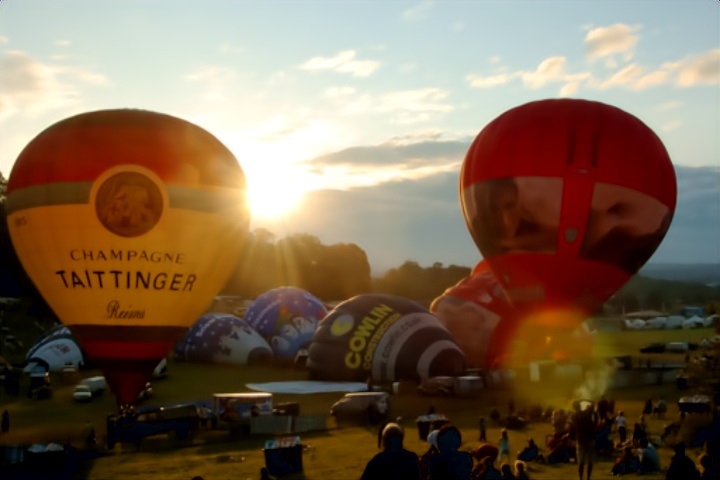}}\
	\subfloat[Zero-DCE]{\includegraphics[width=1.75in]{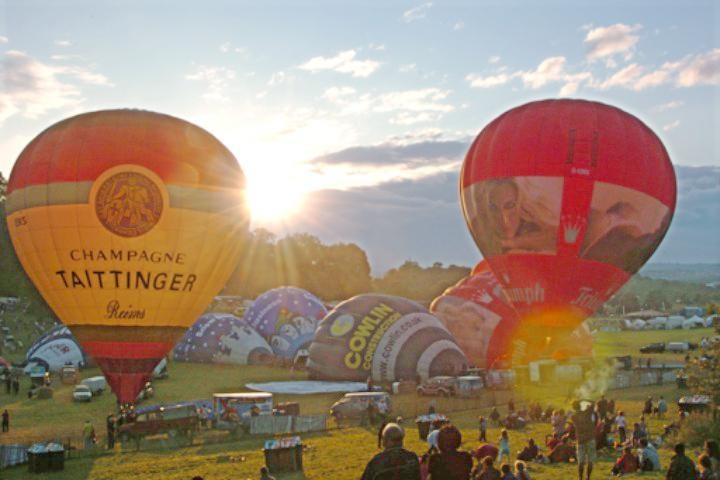}}\
	\subfloat[Ours]{\includegraphics[width=1.75in]{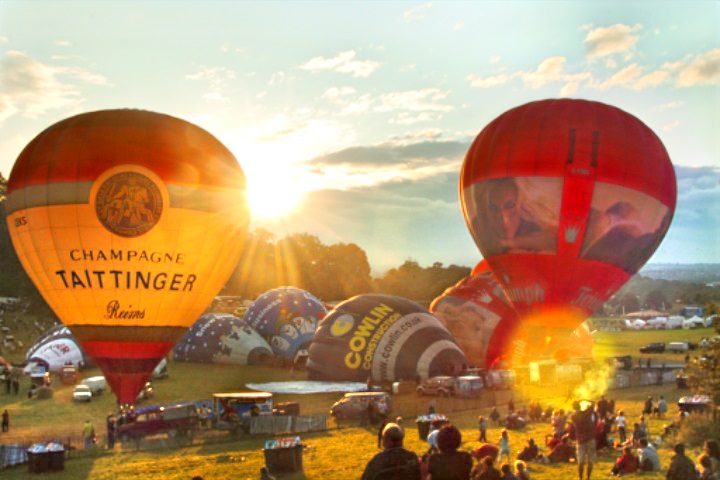}}\vspace{-4mm}
\end{figure*}
\begin{figure*}[!h]
	\centering
	\subfloat[Input]{\includegraphics[width=1.75in]{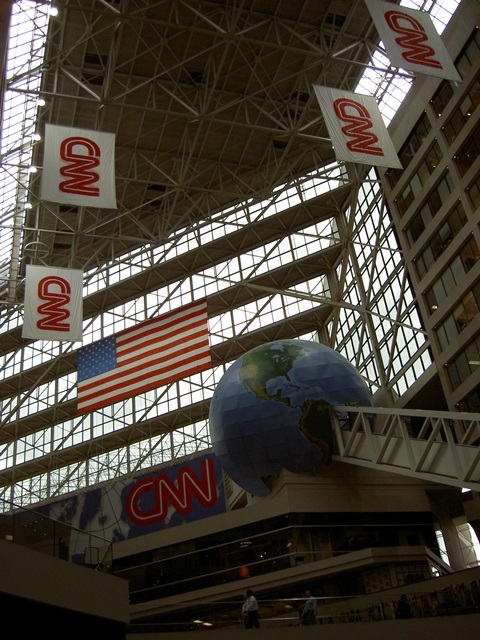}}\
	\subfloat[Dong]{\includegraphics[width=1.75in]{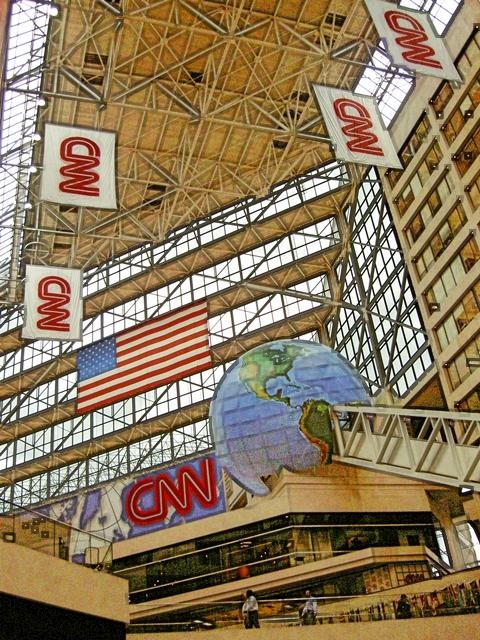}}\
	\subfloat[NPE]{\includegraphics[width=1.75in]{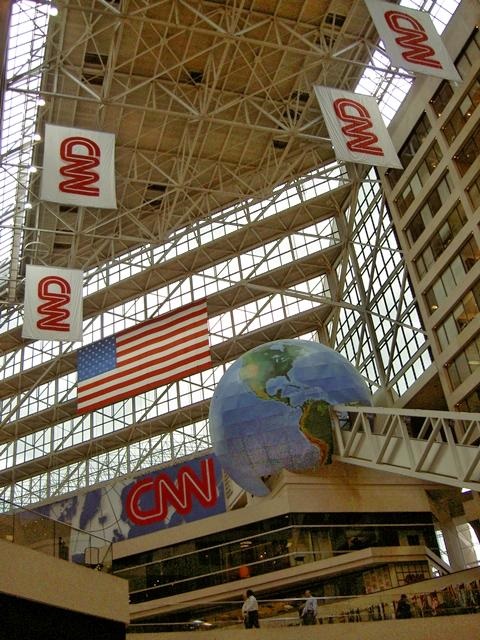}}\
	\subfloat[LIME]{\includegraphics[width=1.75in]{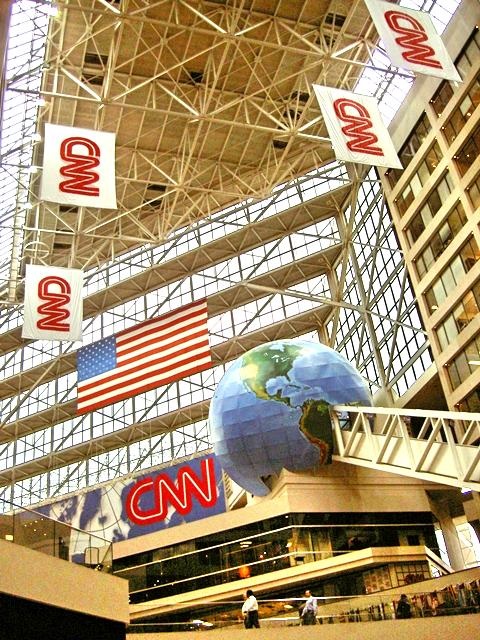}}\\ \vspace{-3mm}
	\subfloat[RetinexNet]{\includegraphics[width=1.75in]{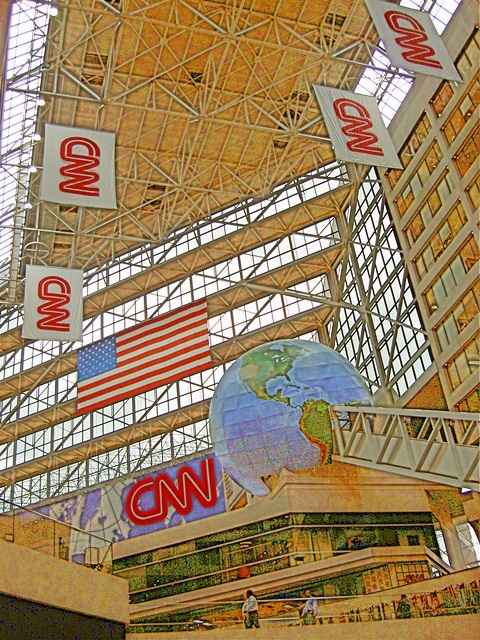}}\
	\subfloat[EG]{\includegraphics[width=1.75in]{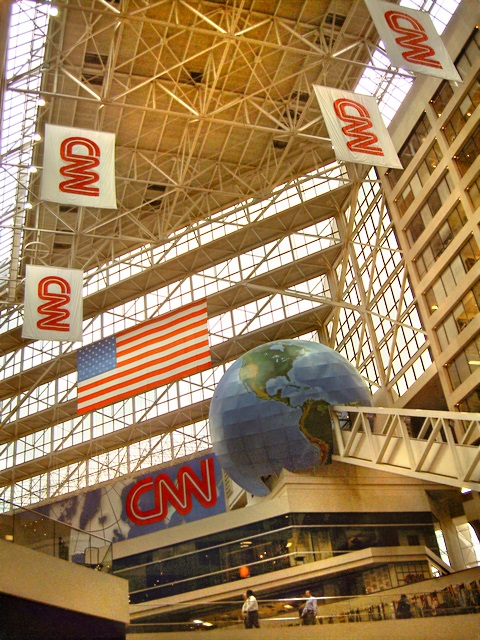}}\
	\subfloat[Zero-DCE]{\includegraphics[width=1.755in]{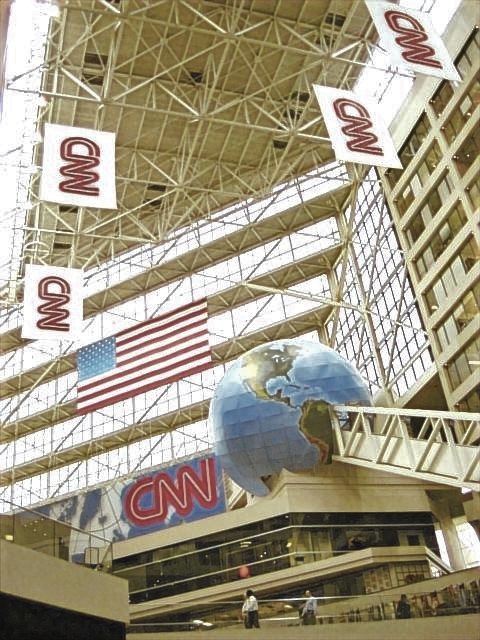}}\
	\subfloat[Ours]{\includegraphics[width=1.75in]{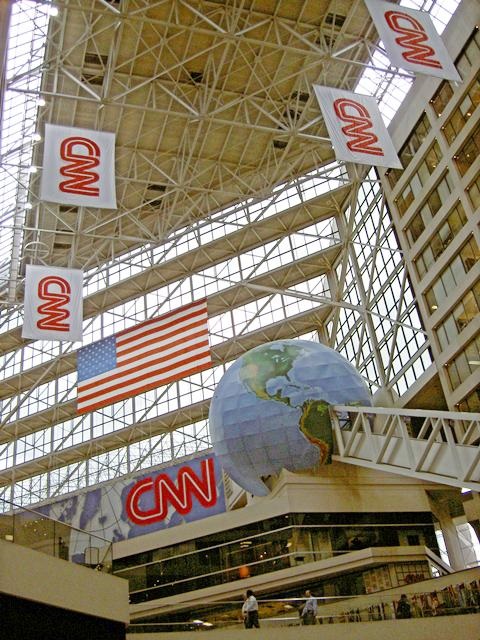}}
	\caption{Visual Comparison with state-of-the-art low-light image enhancement methods on the MEF dataset (row 1-2) and the DICM dataset (row 3-4). Please zoom in for a better review. EG denotes EnlightenGAN.}
	\label{Fig.7}\vspace{-1mm}
\end{figure*}
\begin{table}[!t]
	\renewcommand{\arraystretch}{1.3}
	\caption{Quantitative evaluation of low-light image enhancement methods on the LOL dataset. The best results are highlighted in bold. Note that RetinexNet, DSLR, and MBLLEN are trained on our LSRW dataset.}
	\label{Table II}
	\centering
	\setlength{\tabcolsep}{2.8mm}{
	\begin{tabular}{cccccc}
		\shline
		\textbf{Methods} & \textbf{PSNR}& \textbf{SSIM} &\textbf{FSIM} & \textbf{MAE}&\textbf{GMSD}\\
		\shline
		MF & 16.907 &  0.605 & 0.906 & 0.115 & 0.090 \\
		Dong & 15.905 & 0.537 & 0.875 & 0.142 & 0.119 \\
		NPE & 17.354 & 0.546 & 0.875 & 0.093 & 0.115 \\
		SRIE & 11.557	& 0.531 & 0.887 & 0.220 & 0.103 \\
		BIMEF & 13.769 & 0.640 & 0.907 & 0.103 & 0.085 \\
		MSRCR & 13.964 & 0.514&	0.827 & 0.046 & 0.151\\
		LIME & 17.267& 0.513 & 0.850 & 0.097 & 0.123 \\
		RetinexNet & 16.013 & 0.661 & 0.851 & 0.071 & 0.146 \\
		DSLR & 15.036 & 0.667 & 0.883 & 0.196 & 0.144 \\
		MBLLEN &18.860& 0.754 & 0.904 & \textbf{0.032} & 0.103 \\
		EnlightenGAN & 17.239 & 0.678 & 0.911 & 0.087 & 0.085 \\
		Zero-DCE & 14.584 & 0.610 & 0.911 & 0.161 & 0.087 \\
		Ours & \textbf{20.207} & \textbf{0.816} & \textbf{0.933} & 0.036 & \textbf{0.076} \\
		\shline
	\end{tabular}}
\end{table}

LIME, DICM, NPE, MEF, VV have commonly been used as benchmark datasets for low-light image enhancement methods evaluation, which only contain low-light images so that PSNR and SSIM cannot be used for quantitative evaluation. Therefore, we use non-reference image quality evaluation NIQE to evaluate the performance of our method. The results are shown in Table III. Some visual comparison is illustrated in Fig.7. Please zoom in to compare more details.

\begin{table}[htbp]
	\renewcommand{\arraystretch}{1.3}
	\caption{NIQE scores on the MEF, LIME, NPE, VV, DICM dataset, respectively. The best results are highlighted in bold.}
	\label{Table III}
	\centering
	\setlength{\tabcolsep}{1.8mm}{
	\begin{tabular}{ccccccc}
		\shline
		\textbf{Methods} & \textbf{MEF}& \textbf{LIME} &\textbf{NPE} & \textbf{VV}&\textbf{DICM}&\textbf{AVG}\\
		\shline
		Dong &4.695 &4.052&4.334&3.284&3.263&3.926\\
		NPE &4.256 &3.905&3.403&3.031&2.845&3.488\\
		LIME&4.447&4.155&3.796&2.750&3.001&3.630\\
		RetinexNet&4.408&4.361&3.943&3.816&4.209&4.147\\
		MBLLEN&3.654&4.073&5.000&4.294&3.442&4.063\\
		EnlightenGAN&\textbf{3.573}&3.719&4.113&\textbf{2.581}&3.570&3.443\\
		Zero-DCE&4.024&3.912&3.667&3.217&\textbf{2.835}&3.531\\
		Ours&3.029&\textbf{3.176}&\textbf{3.355}&3.093&3.503&\textbf{3.431}\\
		\shline
	\end{tabular}}
\end{table}

\subsection{User Study}
We conduct a user study to compare the performance of our method and other methods. We collect 20 additional low-light images in the real-world scenes for user study and invite 10 participants to evaluate the enhanced results of the real low-light images obtained using five different methods (NPE, LIME, EnlightenGAN, MBLLEN, and Our method). The participants should consider the contrast, artifacts, noise, details, and color of the enhanced results and rate them according to the performance of the enhanced images (from 1 to 5, 1 means the best, 5 means the worst). Fig.8 shows the distribution of scores and our method gets the best result, which demonstrates that the enhanced images obtained by our method are more visually satisfactory.

\begin{figure}[!h]
	\centering
	\includegraphics[width=3.5in]{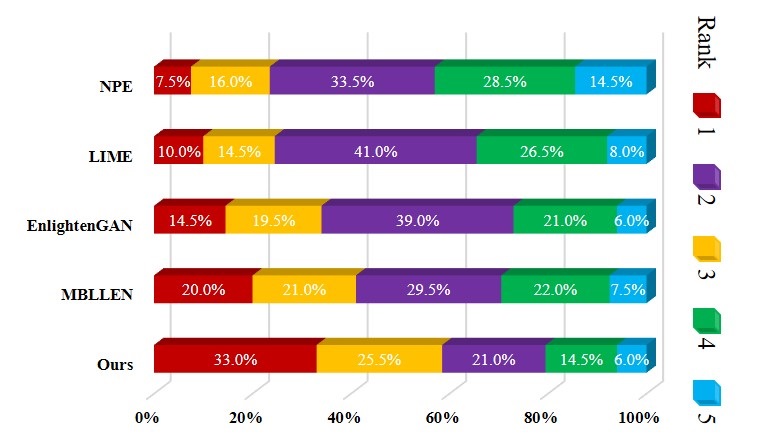}
	\caption{Score distribution of user study. Our method has a better score distribution.}
	\label{Fig.8}\vspace{-1mm}
\end{figure}

\subsection{Ablation Study}
In this section, we quantitatively evaluate the effectiveness of different components and the loss function setting in our model based on the LOL dataset. The results are shown in Table IV.

1). Effectiveness of CEM and DRM: We evaluate the effectiveness of the Contrast Enhancement Module (CEM) and Detail Reconstruction Module (DRM) in the Relight-Net by removing CEM and DRM respectively to build our Relight-Net. Removing CEM or DRM will significantly reduce the performance of our model. As shown in Table IV, the experimental result demonstrated that combing spatial and frequency information can obtain better performance than using one alone.

2). Effectiveness of deep-narrow architecture: We evaluate the effectiveness of DN-ResUnet and compare it with the corresponding "shallow-wide" ResUnet. We replace DN-ResUnet architecture in DenoiseNet and RelightNet with ResUnet. Our proposed DN-ResUnet exceeds ResUnet with 0.971dB (=20.207-19.236) on PSNR and 0.011 (=0.816-0.805) on SSIM. The results demonstrate that our default architecture will result in better performance.

3).	Loss function setting:
In order to explore the effectiveness of the loss function setting, we conduct experiments by converting the content loss into MSE loss, removing the perceptual loss and removing the frequency loss, respectively. Using L1 loss exceeds MSE loss with 0.676dB (=20.207-19.531) on PSNR and 0.012 (=0.816-0.804) on SSIM. Removing the perceptual loss and the frequency loss will lead to performance degradation. After removing the perceptual loss, PSNR decreased by 0.868db (=20.207-19.339) and SSIM decreased by  0.043(=0.816-0.773). After removing the frequency loss, PSNR decreased by 0.451db (=20.207-19.756) and SSIM decreased by 0.012 (=0.816-0.804). The experimental results verify the rationality of our loss function setting. 

\begin{table}[!h]
	\renewcommand{\arraystretch}{1.3}
	\caption{Ablation study. This table reports the performance under each condition based on the LOL dataset. In this table, "w/o" means without.}
	\label{Table IV}
	\centering
	\setlength{\tabcolsep}{5.1mm}{
	\begin{tabular}{ccc}
		\shline
		\textbf{Conditions} & \textbf{PSNR}& \textbf{SSIM}\\
		\shline
		1. default & 20.207 & 0.816\\
		2. w/o CEM & 17.965 & 0.784 \\
		3. w/o DRM & 17.483 & 0.772 \\
		4. DN-ResUnet $\rightarrow$ ResUnet & 19.236 & 0.804 \\
		5. L1 $\rightarrow$ MSE & 19.531 & 0.804 \\
		6. w/o perceptual loss & 19.339 & 0.773 \\
		7. w/o frequency loss& 19.756 & 0.805 \\
		\shline
	\end{tabular}}
\end{table}

\begin{figure*}[!h]
	\centering
	\subfloat[DSFD+Low-light image]{\includegraphics[width=1.75in]{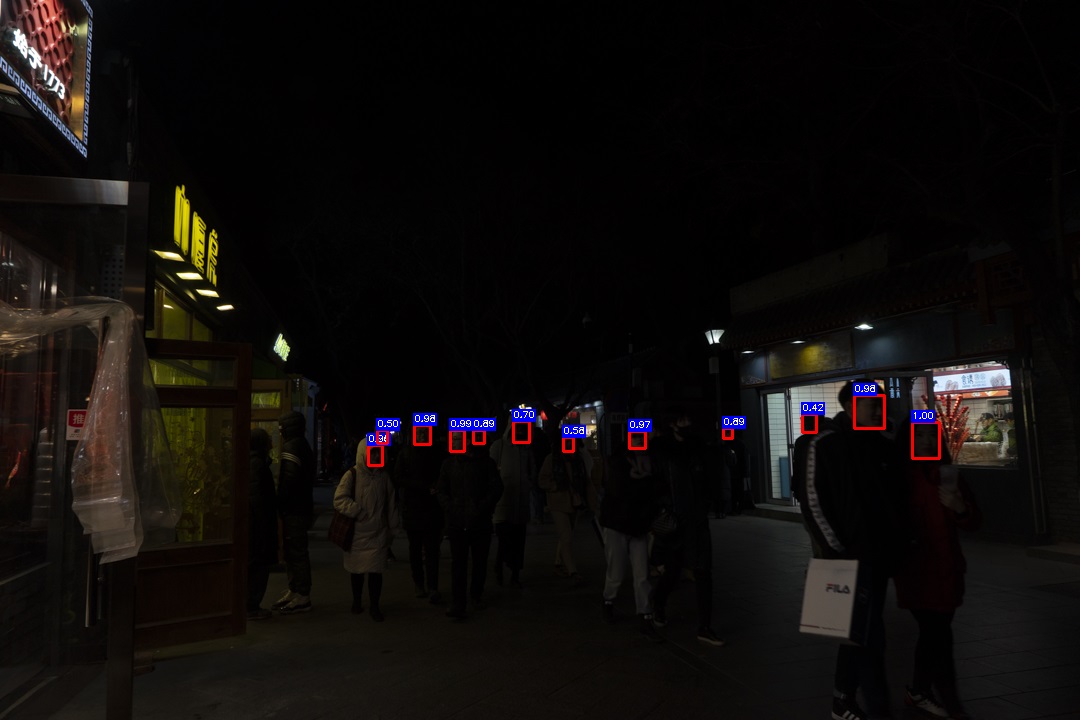}}\
	\subfloat[DSFD+MBLLEN]{\includegraphics[width=1.75in]{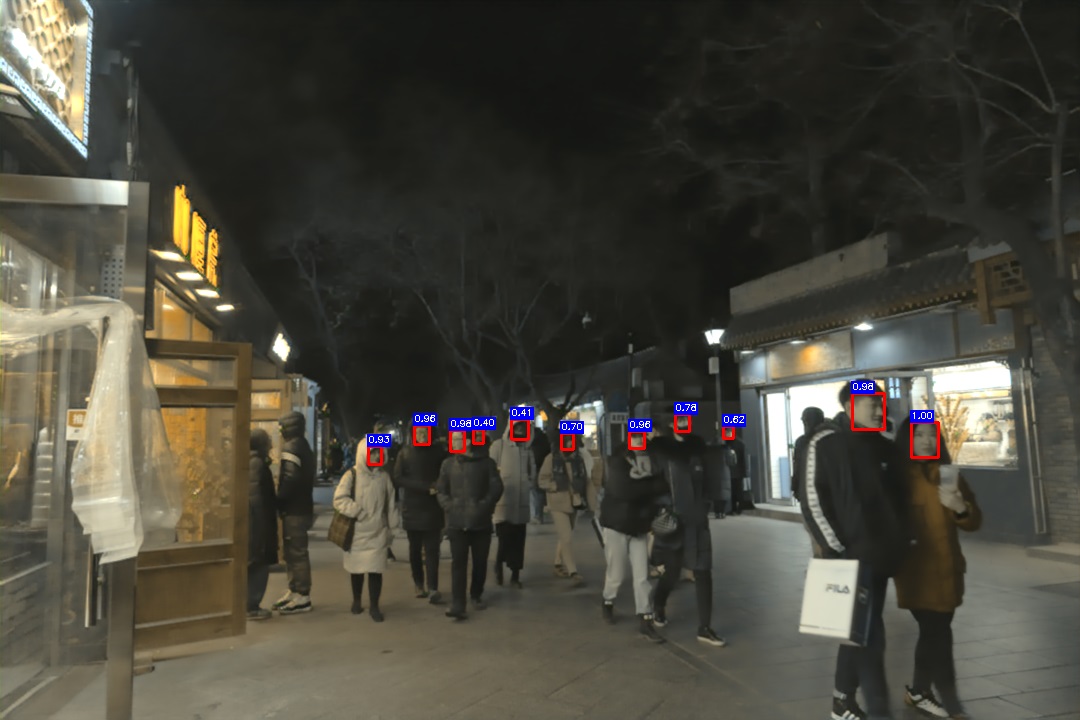}}\
	\subfloat[DSFD+EG]{\includegraphics[width=1.75in]{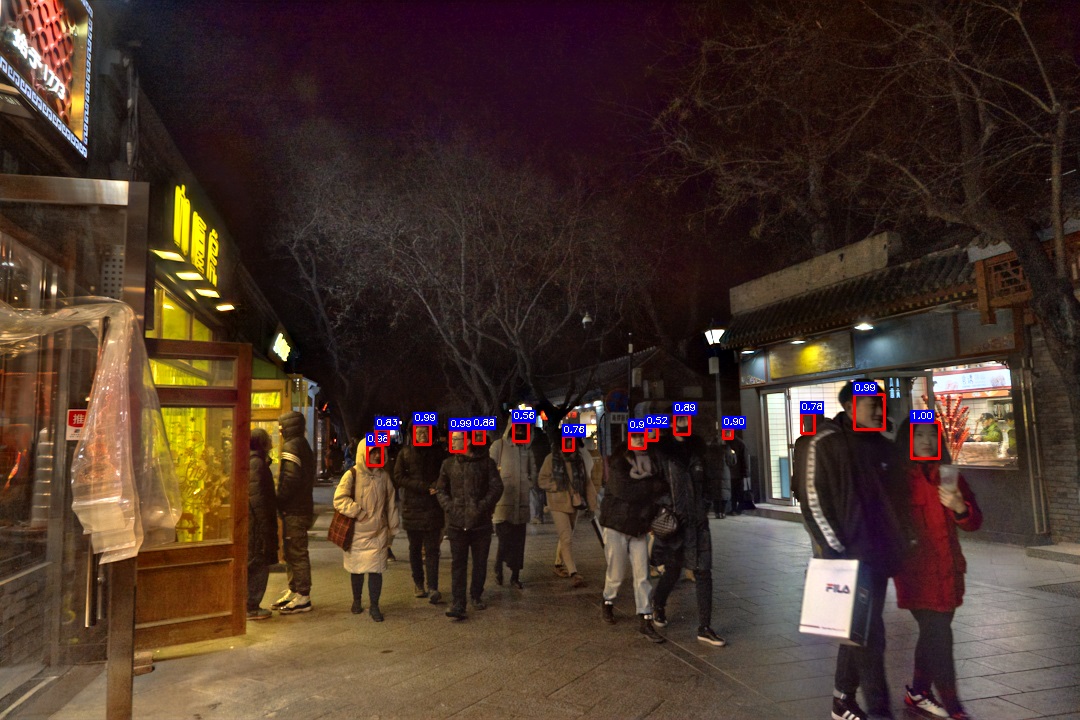}}\
	\subfloat[DSFD+R2RNet]{\includegraphics[width=1.75in]{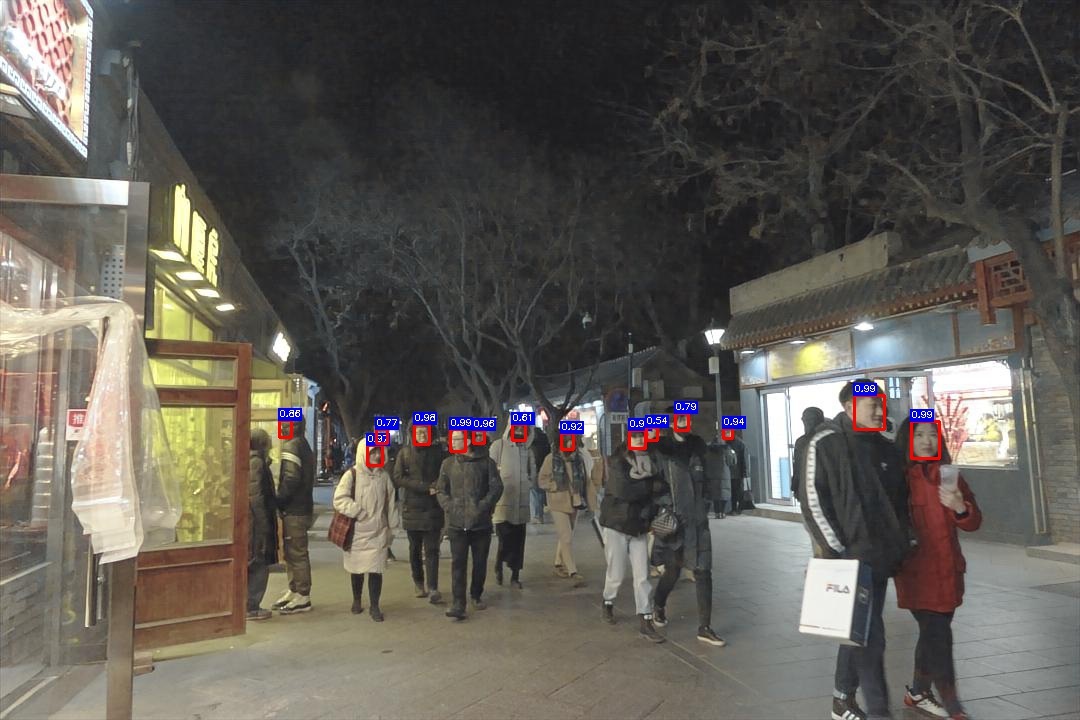}}\vspace{-3mm}
\end{figure*}
\begin{figure*}[!h]
	\centering
	\subfloat[RetinaFace+Low-light image]{\includegraphics[width=1.75in]{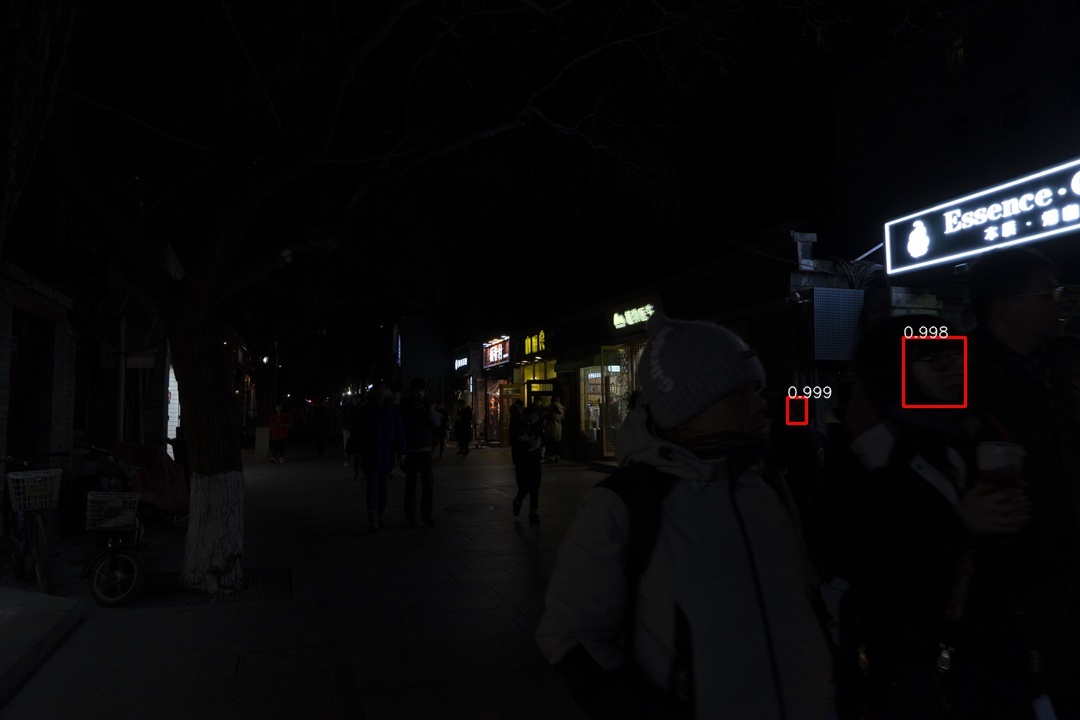}}\
	\subfloat[RetinaFace+MBLLEN]{\includegraphics[width=1.75in]{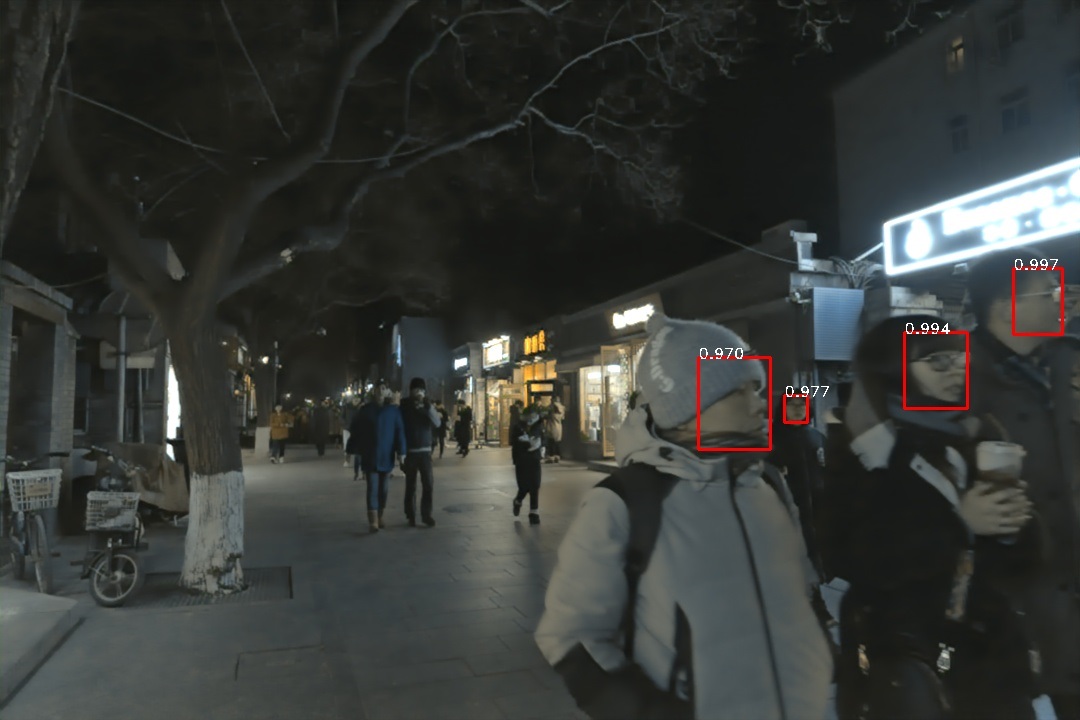}}\
	\subfloat[RetinaFace+EG]{\includegraphics[width=1.75in]{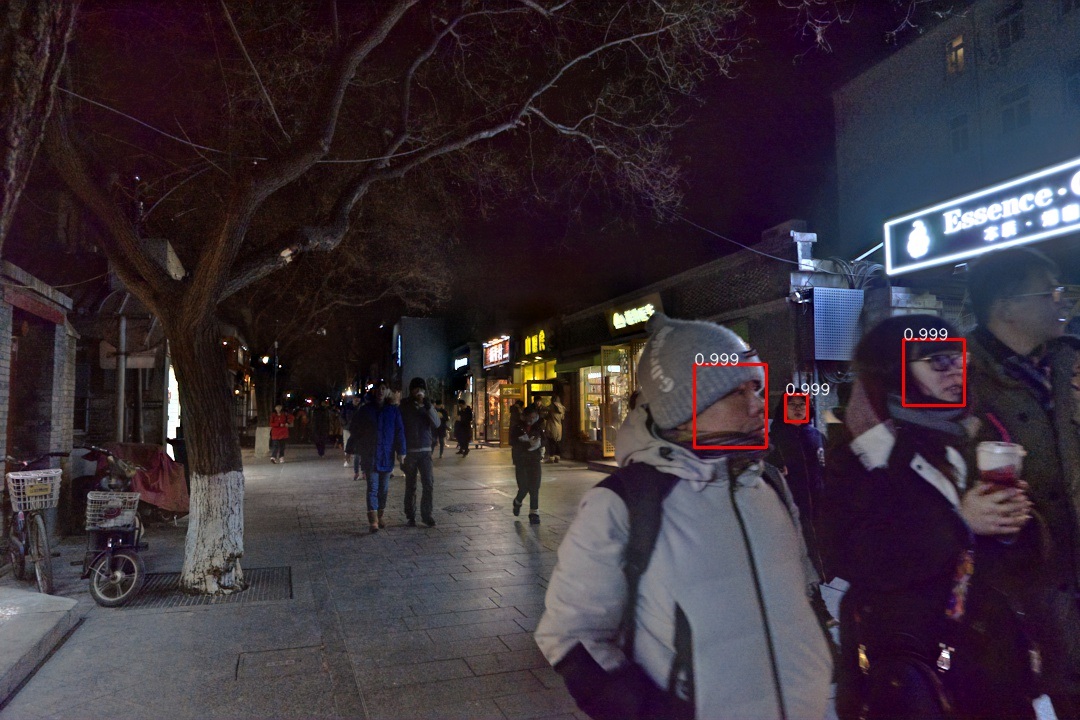}}\
	\subfloat[RetinaFace+R2RNet]{\includegraphics[width=1.75in]{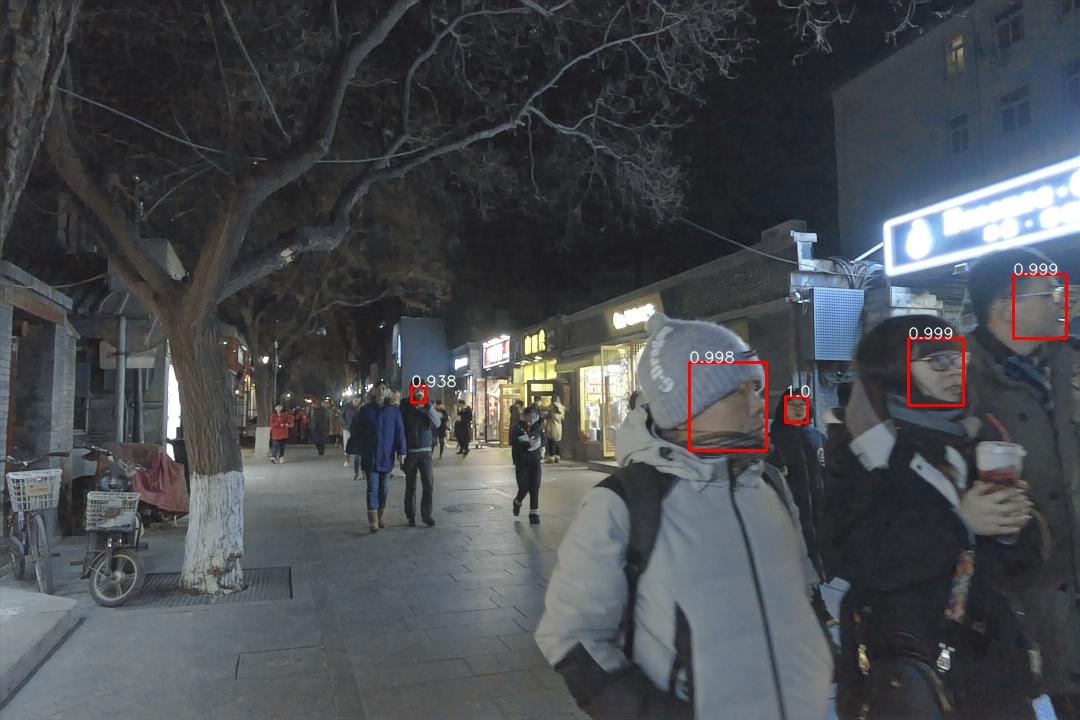}}
	\caption{Examples of face detection rsults. We use EnlightenGAN, MBLLEN, and our R2RNet as pre-processing step, followed by DSFD and RetinaFace for detection. EG denotes EnlightenGAN.}
	\label{Fig.9}\vspace{-1mm}
\end{figure*}
\subsection{Pre-Processing for Improving Face Detection}
Image enhancement as pre-processing for improving subsequent high-level vision tasks has recently received increasing attention \cite{53}, \cite{54}. We investigate the impact of light enhancement on the DARK FACE dataset\footnote{https://flyywh.github.io/CVPRW2019LowLight}, which was specifically built for the task of face detection in low-light conditions. The DARK FACE dataset consists of 6,100 real-world low-light images captured during the nighttime, including 6000 images in the training/validation set, 100 images in the testing set. Because there are no corresponding labels in the test set, we randomly select 100 images from the training set for evaluation, applying our R2RNet as a pre-processing step, followed by two state-of-art pre-trained face detection methods: RetinaFace \cite{55} and DSFD \cite{56}. Using R2RNet as pre-processing improves the average precision (AP) from 17.12$\%$ (DSFD+Low-light image) and 15.28$\%$ (RetinaFace+Low-light image) to 33.98$\%$ (DSFD+R2RNet) and 25.97$\%$ (RetinaFace+R2RNet) after enhancement, which demonstrates that R2RNet can improve the performance of high-level vision tasks, in addition to producing visually pleasing results. We also conduct experiments using EnlightenGAN and MBLLEN. EnlightenGAN improves the AP to 32.75$\%$ and 23.44$\%$, and MBLLEN improves the AP to 31.69$\%$ and 24.67$\%$. Examples of face detection results are illustrated in Fig.9.

\section{Conclusion}
In this research, we proposed a novel Real-low to Real-normal Network for low-light image enhancement based on the Retinex theory, the proposed network includes three sub-networks: a Decom-Net, a Denoise-Net, and a Relight-Net. The enhanced results obtained by our method have better visual quality. Unlike previous methods, we collected the first large-scale real-world paired low/normal-light images dataset known as LSRW dataset used for network training. The results on the publicly available datasets showed that our method can properly improve the image contrast and suppress noise, and achieve the highest PSNR and SSIM scores, which outperform state-of-the-art methods by a large margin. We also showed that our R2RNet can effectively improve the performance of face detection methods under low-light conditions.

Overall, the main contributions of this work are threefold:

1). We proposed a novel Real-low to Real-normal Network (R2RNet) to transform the weakly illuminated image into the normal-light image. The proposed network consists of three subnets: a Decom-Net, a Denosie-Net, and a Relight-Net. The purpose of Decom-Net is to decompose the input image into an illumination map and a reflectance map based on the Retinex theory. The Denosie-Net is designed to suppress noise in the reflectance map. The Relight-Net uses spatial information of low-light images to improve contrast and uses frequency information for detail reconstruction. Additionally, a novel frequency loss function was used to help Relight-Net recover more image details.

2). Different from previous methods using synthetic image datasets, we collected the first large-scale real-world paired low/normal-light image dataset (LSRW dataset), which contains 5650 pairs of low/normal-light images to satisfy the training requirement of deep neural networks and make our model has better generalization performance in real-world scenes.

3). The experimental results on the publicly available datasets demonstrated that our method outperforms state-of-the-art methods by a large margin. The enhanced results generated by our method are excellent in contrast, brightness, detail preservation, and noise suppression. And we also showed that our method can effectively improve the performance of face detection under insufficient illumination conditions.

In the future, we will explore a more effective model and apply the model to other enhancement tasks (such as low-light video enhancement, extremely low-light image enhancement, \emph{etc}.).

\section*{Acknowledgment}
We would like to thank Mingrui Wu, Zhiyun Jiang, and Wenxuan Liu for helping us collect the LSRW dataset.

\ifCLASSOPTIONcaptionsoff
  \newpage
\fi
\bibliographystyle{IEEEtran}
\bibliography{references}

\end{document}